\documentclass[lettersize,journal]{IEEEtran}
\usepackage{amsmath,amsfonts}
\usepackage{algorithmic}
\usepackage{algorithm}
\usepackage{array}
\usepackage{subfigure}
\usepackage{textcomp}
\usepackage{stfloats}
\usepackage{url}
\usepackage{verbatim}
\usepackage{graphicx}
\usepackage{cite}
\usepackage{xcolor}
\usepackage{subfiles}
\usepackage{multirow}
\usepackage{amsmath}
\usepackage{subcaption}
\usepackage{soul}
\usepackage{mathtools}

\begin{document}

\title{Prediction of Lane Change Intentions of Human Drivers using an LSTM, a CNN and a Transformer}

\author{Francesco De Cristofaro, Felix Hofbaur, Aixi Yang, Arno Eichberger
\thanks{This work has been submitted to the IEEE for possible publication. Copyright may be transferred without notice, after which this version may no longer be accessible.}
\thanks{}}

\markboth{}%
{Shell \MakeLowercase{\textit{et al.}}: A Sample Article Using IEEEtran.cls for IEEE Journals}


\maketitle

\begin{abstract}
Lane changes of preceding vehicles have a great impact on the motion planning of automated vehicles especially in complex traffic situations. Predicting them would benefit the public in terms of safety and efficiency. While many research efforts have been made in this direction, few concentrated on predicting maneuvers within a set time interval compared to predicting at a set prediction time. In addition, there exist a lack of comparisons between different architectures to try to determine the best performing one and to assess how to correctly choose the input for such models. In this paper the structure of an LSTM, a CNN and a Transformer network are described and implemented to predict the intention of human drivers to perform a lane change. We show how the data was prepared starting from a publicly available dataset (highD), which features were used, how the networks were designed and finally we compare the results of the three networks with different configurations of input data. We found that transformer networks performed better than the other networks and was less affected by overfitting. The accuracy of the method spanned from $82.79\%$ to $96.73\%$ for different input configurations and showed overall good performances considering also precision and recall.
\end{abstract}


\section{Introduction}\label{sec:Intro}
Lane changes constitute a challenging task for drivers, both for human and automated. In complex traffic scenarios the behavior of surrounding vehicles related to lane change influence comfort and safety of the other traffic participants. In the scope of automated driving, cut ins and cut outs of surrounding vehicles may influence the trajectory planner module of the driving controller leading to sudden decelerations, time loss and a general discomfort for the passengers. In extreme cases, such behaviors could lead to safety risks for the passengers. To avoid such consequences a possible solution is to implement a prediction algorithm within the planning module to timely identify surrounding vehicles which will perform a lane change and modify the driving strategy. 

The interest in lane change intention prediction among researchers in the automotive field has increased steadily in the last years. While in most of the publications authors tried to predict lane change maneuvers with a preset prediction time (i.e. the time passing between the prediction and the execution of the maneuver) or provided an average of their achieved prediction times  (e.g. \cite{ref13}, \cite{ref16}, \cite{ref24}, \cite{ref26}, \cite{ref50}, \cite{ref84}, \cite{ref93}, \cite{ref159}), a few others focused on predicting lane changes within a future time interval (e.g. \cite{ref_Schlechtriemen}). While the former approach leads to clear results in which the predictive power of the proposed method can be expressed through the preset (or average) prediction time, the latter might reveal itself more versatile. In fact a network which is trained to recognize lane changes at a specific instant before they happen is not guaranteed to identify said lane changes before or after that instant.

In this paper we utilize the highD dataset \cite{ref_highD} to train an LSTM, a CNN and a Transformer network to predict lane change maneuvers within a future time interval. While neither of these networks are new to being utilized to predict lane change maneuvers (e.g. \cite{ref16}, \cite{ref81}, \cite{ref126}), we believe that there is still a scarcity of comparisons between them to access which one is more suitable for this task and that there is not enough analysis on how these networks behave when the task is to predict within a time interval and not at a specific instant. In addition, we will also discuss the effects of data processing on the results, which is an aspect often overlooked in the available literature.

The paper is structured as follows: in Section \ref{sec:data} the problem is described and highD dataset is presented and briefly discussed in addition to an explanation on data labeling and preparation. In Section \ref{sec:methodology} LSTMs, CNNs and Transformers are introduced and the task of designing them is described. In Section \ref{sec:results} the experiments are explained and the results of the prediction task are presented and discussed. Finally, Section \ref{sec:conclusion} contains our final comments and recommendations for future developments of the research.

\section{Problem Definition and Input Data}\label{sec:data}
Although highway scenarios have already been covered by many researchers, only recently authors opted to use a modern and publicly available dataset such as highD. To make our work more easily repeatable and comparable with future publications we also opted to use the highD dataset.

HighD dataset is a dataset of naturalistic driving trajectory collected on German highways in the years 2017-2018 using drones at a frequency of $25$Hz \cite{ref_highD}. The whole dataset includes $147$ hours of driving for a total of more than $110,000$ vehicle trajectories. Before proceeding with the processing and labeling of the dataset it is important to understand which scenario is considered in this work, which problem is tackled and how data is used to solve it. In this section these themes will be dealt with and the data preparation will be explained in detail.

\subsection{Scenario Definition}
\begin{figure}
    \centering
    \includegraphics[width=0.9\linewidth]{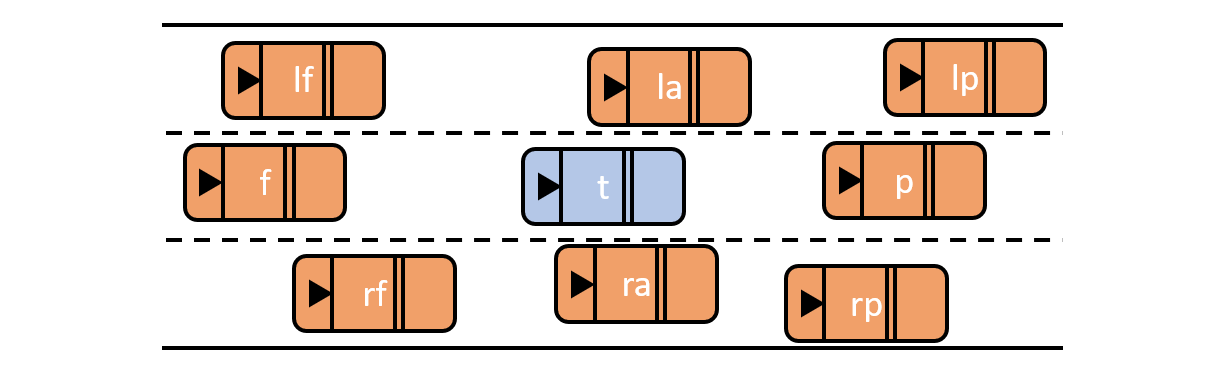}
    \caption{Scenario considered in this work. The target (t) vehicle is surrounded by the right following vehicle (rf), the right alongside vehicle (ra), the right preceding vehicle (rp), the following alongside vehicle (fa), the preceding vehicle (p), the left following vehicle (lf), the left alongside vehicle (la) and the left preceding vehicle (lp).}
    \label{fig:scenario}
\end{figure}

The chosen scenario is a highway scenario. The focus will be on predicting the behavior of a single vehicle (called target vehicle) and in doing so its surrounding environment will also be taken in consideration, see Fig. \ref{fig:scenario}. In particular, for each target vehicle a maximum of eight surrounding vehicles will be identified.

\subsection{Problem Definition}
The goal is to predict if the target vehicle will perform a lane change maneuver (LC) within the next $\Delta t_{p,\textnormal{MAX}}$ seconds (maximum prediction time) or if it will perform a lane keeping maneuver (LK). In addition, it would be also interesting to be able to tell if the LC will be a left lane change (LLC) or a right lane change (RLC). The problem is hence defined as a multi-classification problem with three output classes. To predict a LC, the last $\Delta t_o$ seconds (observation window) of the trajectory of the target vehicle (the vehicle on which the prediction will be 
made) are observed to make a prediction based on them. 

In order to train a machine learning (ML) model to be able to perform such prediction it is necessary to prepare a number of trajectories of uniform length extracted from the highD dataset, label them according if they precede a LK, a LLC or a RLC and use them to train and test said method.

\subsection{Data cutting and labeling}
HighD dataset is structured as a sequence of trajectories, each trajectory identified by an id and referencing the id's of the surrounding vehicles. For now let's assume that $\Delta t_{p,\textnormal{MAX}}$ and $\Delta t_o$ have been set.

\begin{figure}
    \centering
    \includegraphics[width=1\linewidth]{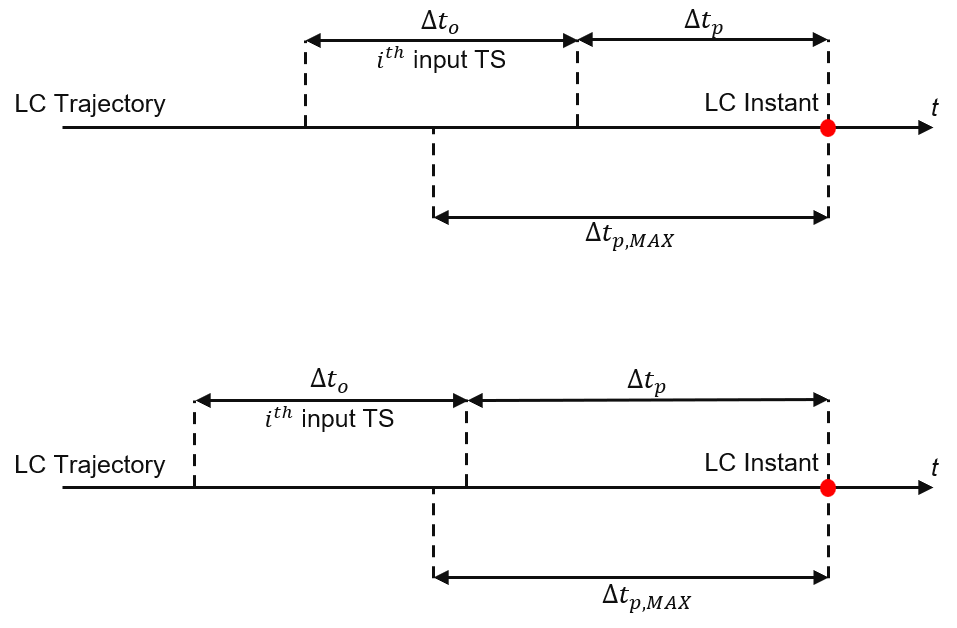}
    \caption{Two possible ways of cutting a LC input TS from a trajectory which features a LC (LC trajectory). The prediction time must be $0 < \Delta t_{p} < \Delta t_{p,\textnormal{MAX}}$.}
    \label{fig:LC_cases}
\end{figure}

\begin{figure}
    \centering
    \includegraphics[width=1\linewidth]{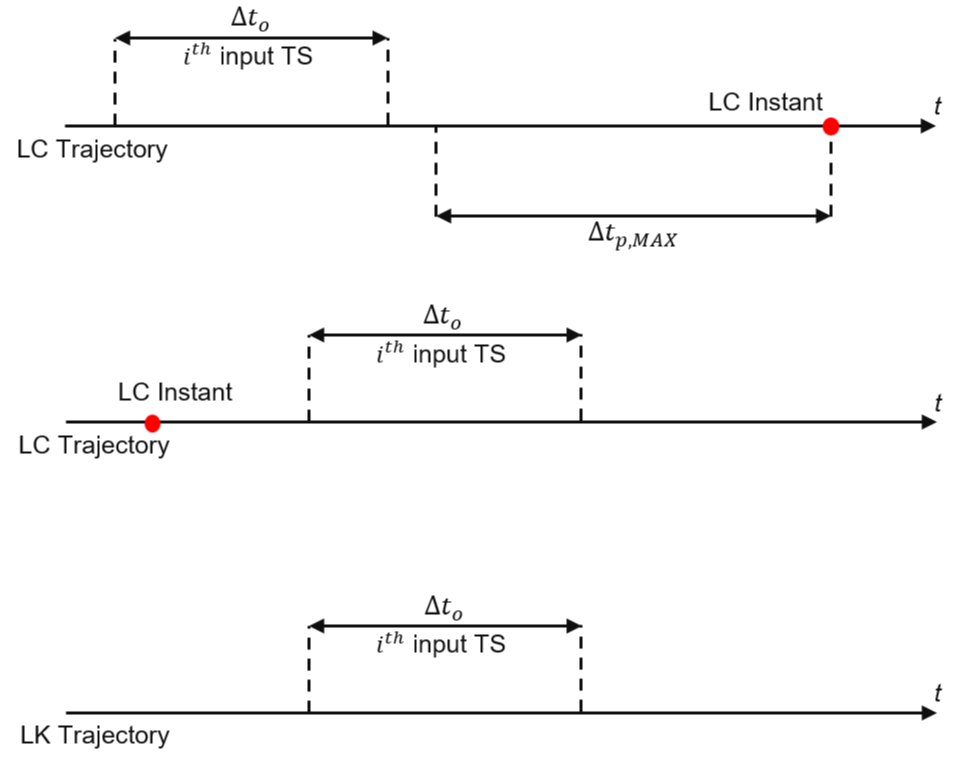}
    \caption{Three possible ways of cutting a LK input TS. An input TS is labeled as a LK if the ending instant of the TS takes place earlier than $\Delta t_{p,\textnormal{MAX}}$ from a LC instant, if it takes place after a LC instant or if is cut from a trajectory which does not feature a LC. Input TSs of length $\Delta t_o$ which start earlier than a LC instant and end after it are excluded to avoid reusing sections of the same trajectory in the processed dataset (there would be overlap with the LC input TSs).}
    \label{fig:LK_cases}
\end{figure}

Firstly, LC instants are identified for each trajectory. The LC instant is defined as the instant in which the lane id of a trajectory changes in the highD dataset which corresponds to the instant in which the center of the vehicle crosses the lane line (the most common definition). For each LC instant a LC trajectory segment of length $\Delta t_o$ is selected with a prediction time $\Delta t_p$ extracted with a uniform distribution between $0\textnormal{s}$ and $\Delta t_{p,\textnormal{MAX}}$. If the length of a trajectory before a LC instant is shorter than $\Delta t_o + \Delta t_{p,\textnormal{MAX}}$, no LC trajectory segment is selected for this LC instant. If a LC trajectory segment contains another LC instant it is discarded for simplicity of data handling. Each LC trajectory segment is labeled with either LLC or RLC depending on if the LC instant which they precede is a LLC or a RLC. Then, for each trajectory in the highD dataset, a single LK trajectory segment of length $\Delta t_o$ will be selected. A LK trajectory segment is defined as a trajectory segment which does not contain any LC instant and whose ending instant does not precede a LC instant by a time between $0\textnormal{s}$ and $\Delta t_{p,\textnormal{MAX}}$. If no LK trajectory segment can be found within a trajectory in the highD dataset, no LK trajectory segment will be selected in that trajectory. If multiple LK trajectory segments are available for a specific trajectory in the highD dataset, only one is chosen randomly and selected. All LK trajectory segments are labeled as LK. The possible ways in which a LC input time series (TS) is defined are shown in Fig. \ref{fig:LC_cases}. The possible ways in which a LK input TS is defined are shown in Fig. \ref{fig:LK_cases}. A maximum of one LK input TS is extracted from a single trajectory in the highD dataset to avoid reusing data points multiple times.

Finally, all the selected and labeled LC and LK trajectory segments are joined and, after being processed to select the desired features, will constitute the dataset used for training and testing.



\subsection{Feature extraction}
The original trajectories from highD include many features (position, velocity, acceleration, lane id etc.) but not all of these are equally important or necessary to predict LC intentions. Moreover, using all the features would increase the complexity of the problem and make the training and testing of the networks slower and heavier. We opted to utilize a subset of the input features and a description of this subset is shown in Tab. \ref{tab:input}.

\begin{table}[]
    \centering
    \begin{tabular}{c}
     \textbf{Input feature description} \\ \hline
     Lateral position of the target vehicle \\ \hline 
     Longitudinal position of the target vehicle\\ \hline
     Lateral velocity of the target vehicle \\ \hline 
     Longitudinal velocity of the target vehicle \\ \hline
     Lateral distance of the surrounding vehicles to the target vehicle\\ \hline
     Longitudinal distance of the surrounding vehicles\\  
     to the target vehicle\\  \hline
     Lateral velocity of the surrounding vehicles to the target vehicle\\ \hline
     Longitudinal velocity of the surrounding vehicles to the target vehicle\\
     \\
     \\
    \end{tabular}
    \caption{Input features}
    \label{tab:input}
\end{table}

\begin{figure}
    \centering
    \includegraphics[width=1\linewidth]{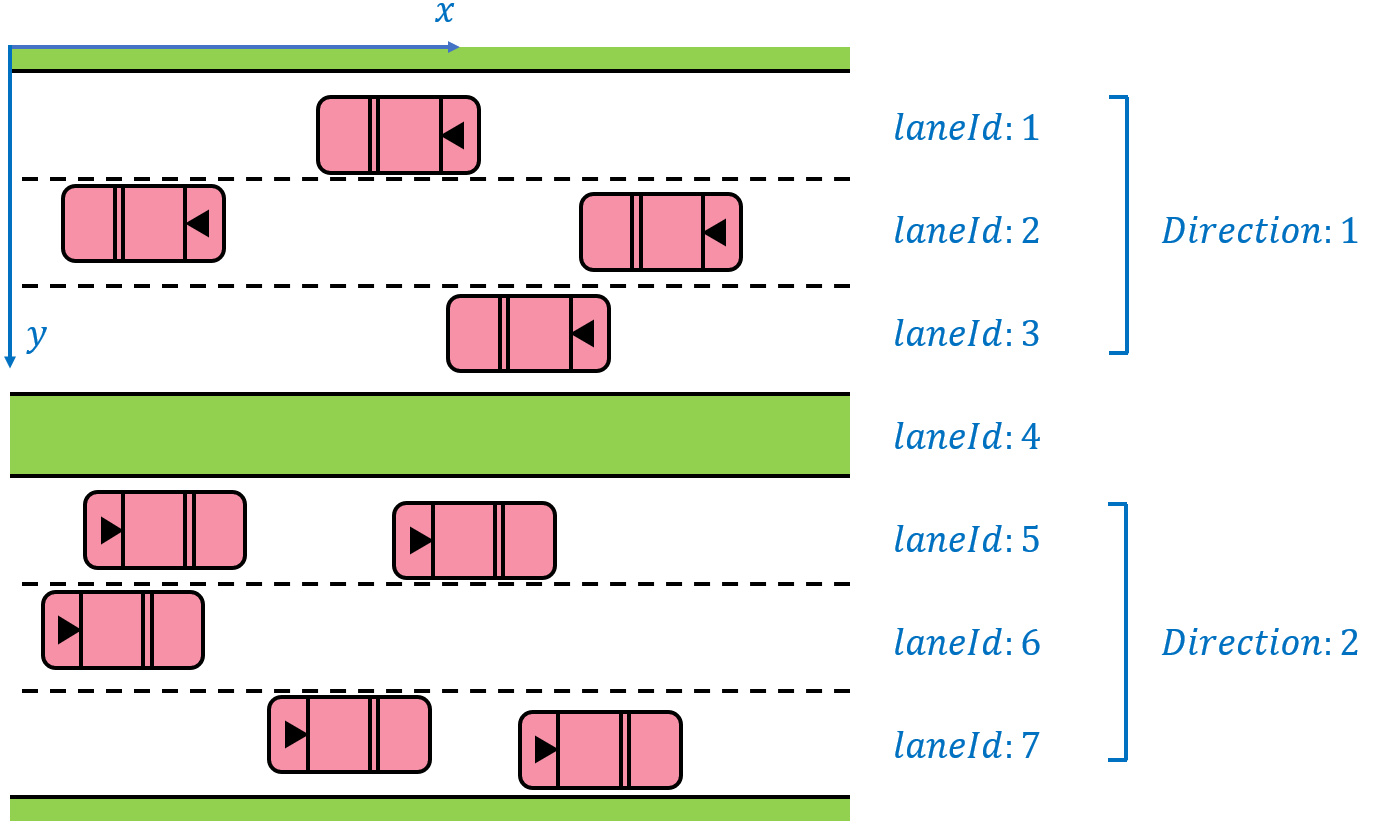}
    \caption{Image coordinate system as defined in the highD dataset in the case of a highway with three lanes per driving direction. Lane numeration and direction identifiers are also shown.}
    \label{fig:coordinates}
\end{figure}

The coordinate system of the highD datasets is the so called image coordinate system. With respect to the images included in the dataset, the x axis runs parallel to the the road while the y axis is perpendicular to it and pointing downward. The image coordinate system as defined in the highD dataset is shown in Fig. \ref{fig:coordinates}. The different driving directions of the vehicles were accounted for by inverting the sign of the features when needed. The calculation of the longitudinal and lateral positions of the target vehicle $x_t$ and $y_t$ at a specific instant $i$ is:
\begin{gather}
 x_t^i = \begin{cases}
    -x_{t,\textnormal{hD}}^i\;\;if\;\;direction=1\;\;\;\;\;\;\;\;\\
        x_{t,\textnormal{hD}}^i\;\;else
    \end{cases}\\
    y_t^i = \begin{cases}
    y_{t,\textnormal{hD}}^i\;\;if\;\;direction=1\;\;\;\;\;\;\;\;\\
        -y_{t,\textnormal{hD}}^i\;\;else
    \end{cases} 
\end{gather}
Where $x_{t,\textnormal{hD}}^i$ and $y_{t,\textnormal{hD}}^i$ are the longitudinal and lateral position of the target vehicle in the highD dataset at the instant $i$.

The calculation of the longitudinal and lateral velocities of the target vehicle $v_{x,t}$ and $v_{y,t}$ at a specific instant $i$ is:
\begin{gather}
 v_{x,t}^i = \begin{cases}
    -v_{x,t,\textnormal{hD}}^i\;\;if\;\;direction=1\;\;\;\;\;\;\;\;\\
        v_{x,t,\textnormal{hD}}^i\;\;else
    \end{cases}\\
    v_{y,t}^i = \begin{cases}
    v_{y,t,\textnormal{hD}}^i\;\;if\;\;direction=1\;\;\;\;\;\;\;\;\\
        -v_{y,t,\textnormal{hD}}^i\;\;else
    \end{cases} 
\end{gather}
where $v_{x,t,\textnormal{hD}}^i$ and $v_{y,t,\textnormal{hD}}^i$ are the longitudinal and lateral velocities of the target vehicle in the highD dataset at the instant $i$.

For the calculations of the distances and velocities of the surrounding traffic vehicles only the calculations for the left preceding vehicle will be shown. For all the other vehicles the calculations are analogous. The calculations of the longitudinal and lateral distances of the left preceding vehicle with respect to the target vehicle at the instant $i$ ($\Delta x_{lp}^i$ and $\Delta y_{lp}^i$ respectively) are:
\begin{gather}
    \Delta x_{lp}^i = \begin{cases}
        x_{t,\textnormal{hD}}^i-x_{lp,\textnormal{hD}}^i\;\;if\;\;direction=1\\
        x_{lp,\textnormal{hD}}^i-x_{t,\textnormal{hD}}^i\;\;else
    \end{cases}\\
    \Delta y_{lp}^i = \begin{cases}
        y_{lp,\textnormal{hD}}^i-y_{t,\textnormal{hD}}^i\;\;if\;\;direction=1\\
        y_{t,\textnormal{hD}}^i-y_{lp,\textnormal{hD}}^i\;\;else
    \end{cases} 
\end{gather}
where $x_{lp,\textnormal{hD}}^i$ and $y_{lp,\textnormal{hD}}^i$ are respectively the longitudinal and lateral position of the left preceding vehicle in the highD dataset at the instant $i$.

The calculations of the longitudinal and lateral velocities of the left preceding vehicle at the instant $i$ ($v_{x,lp}^i$ and $\Delta v_{y,lp}^i$ respectively) are:
\begin{gather}
    v_{x,lp}^i = \begin{cases}
        -v_{x,lp,\textnormal{hD}}^i\;\;if\;\;direction=1\\
        v_{x,lp,\textnormal{hD}}^i\;\;else
    \end{cases}\\
    v_{y,lp}^i = \begin{cases}
        v_{y,lp,\textnormal{hD}}^i\;\;if\;\;direction=1\\
        -v_{y,lp,\textnormal{hD}}^i\;\;else
    \end{cases} 
\end{gather}
where $v_{x,lp,\textnormal{hD}}^i$ and $v_{y,lp,\textnormal{hD}}^i$ are respectively the longitudinal and lateral velocities of the left preceding vehicle in the highD dataset at the instant $i$.

Finally, each sample of the resulting dataset (which will later be used to train and test the networks) will be composed of an input multivariate time series, or trajectory sample, $\overline{X}$ and its label $\overline{y}$ defined as:
\begin{gather}
    \overline{X} \in \mathbb{R}^{n \times d}\\
    \overline{y} \in \{LK,LLC,RLC\}
\end{gather}
where $n=\Delta t_o f_{\textnormal{hD}}$ ($f_{\textnormal{hD}}$ is the frequency of the trajectories in the highD dataset) and d is the number of features (36 in our case). Each row $\overline{x}_j$ of $\overline{X}$ is a vector $\overline{x}_j \in \mathbb{R}^d$ with $j = 1,...,n $ defined as:

\begin{gather}
    \overline{x}_j = [y_t^j, x_t^j, v_{y,t}^j, v_{x,t}^j, \Delta y_{p}^j, \Delta x_{p}^j, v_{y,p}^j, v_{x,p}^j, \notag\\
    \Delta y_{f}^j, \Delta x_{f}^j, v_{y,f}^j, v_{x,f}^j,  \Delta x_{lp}^j \Delta y_{lp}^j, v_{y,lp}^j, v_{x,lp}^j, \notag\\  \Delta x_{la}^j, \Delta y_{la}^j, v_{y,la}^j, v_{x,la}^j,  \Delta x_{lf}^j, \Delta y_{lf}^j, v_{y,lf}^j, v_{x,lf}^j, \notag\\ \Delta x_{rp}^j, \Delta y_{rp}^j, v_{y,rp}^j, v_{x,rp}^j,  \Delta x_{ra}^j, \Delta y_{ra}^j, v_{y,ra}^j, \notag\\ v_{x,ra}^j,  \Delta x_{rf}^j, \Delta y_{rf}^j, v_{y,rf}^j, v_{x,rf}^j]
\end{gather}

where the pedicels (p, f, lp etc.) indicate the surrounding vehicles (see Fig. \ref{fig:scenario}). The $j^{th}$ time-step of the input trajectory $\overline{X}$ will be referred to as $\overline{x}_j$.

\subsection{Dataset parameters configurations}
It was assumed earlier that both $\Delta t_o$'s and $\Delta t_{p,\textnormal{MAX}}$'s values had already been assigned but it was not discuss how. These parameters are expected to significantly influence the performance of the algorithms by determining the size of the input and the maximum preview, i.e. the period of time before a LC instant in which the algorithm is expected to correctly identify the upcoming maneuver. For this reason, it is important to properly select them. To identify the best configuration for these parameters, twelve combinations of values are proposed and tested with varying values of $\Delta t_o$ between $1\textnormal{s}$ and $3\textnormal{s}$ and $\Delta t_{p,\textnormal{MAX}}$ between $3\textnormal{s}$ and $6\textnormal{s}$. The different configurations and the number of maneuvers for each combination can be found in Tab. \ref{tab:configurations}. For each of these configurations the number of LK samples in the data is superior to the number of LLC and RLC samples but the number of LK samples used for training, validating and testing is set to the sum of the number of LLCs and RLCs to avoid unbalanced datasets (the set number of LKs is extracted randomly with a fixed seed). The number of segments per class will change for different configurations of $\Delta t_o$ and $\Delta t_{p,\textnormal{MAX}}$ since some LC instant may not have enough preceding data points depending on the configuration. For each combination of $\Delta t_{p,\textnormal{MAX}}$ and $\Delta t_o$ and for each of the tested architectures a fixed seed was specified when splitting the data in training, validation and test datasets. $60\%$ of the samples were allocated for training, $20\%$ for validation and $20\%$ for testing. The fixed seeds when reducing the number of LK samples and when splitting the data improved consistency and repeatability and allowed us to perform meaningful comparisons between different algorithms since they are trained and tested on the same data.

\begin{table*}[]
    \centering
    \begin{tabular}{c|c|c|c|c|}
         \multicolumn{2}{c}{} & \multicolumn{3}{c}{$\Delta t_o$}  \\ \cline{3-5}
         \multicolumn{2}{c|}{} & $1\textnormal{s}$ & $2\textnormal{s}$ & $3\textnormal{s}$ \\ \cline{2-5}
         \multirow{4}{*}{$\Delta t_{p,\textnormal{MAX}}$} & $3\textnormal{s}$ & (LK: 9517, LLC: 4276, RLC: 5241) & (LK: 8412, LLC: 3773, RLC: 4639) & (LK: 7261, LLC: 3293, RLC: 3968) \\ \cline{2-5}
         &$4\textnormal{s}$&(LK: 8412, LLC: 3773, RLC: 4639)&(LK: 7261, LLC: 3298, RLC: 3968)&(LK: 6099, LLC: 2781, RLC: 3318) \\ \cline{2-5}
         &$5\textnormal{s}$&(LK: 7261, LLC: 3293, RLC: 3968)&(LK: 6099, LLC: 2781, RLC: 3318)&(LK: 4927, LLC: 2299, RLC: 2628) \\ \cline{2-5}
         &$6\textnormal{s}$&(LK: 6099, LLC: 2781, RLC: 3318)&(LK: 4927, LLC: 2299, RLC: 2628)&(LK: 3835, LLC: 1820, RLC: 2015) \\ \cline{2-5}
         \multicolumn{5}{c}{}\\
         \multicolumn{5}{c}{}\\
         
    \end{tabular}
    \caption{The nine different configurations of ($\Delta t_o$, $\Delta t_{p,\textnormal{MAX}}$) and the respective number of samples per output class.}
    \label{tab:configurations}
\end{table*}

\section{Methodology}\label{sec:methodology}
In this section the three considered machine learning approaches to predict LC intentions of human drivers are analyzed. For each of them an introduction to the general architecture is given followed by the specific structure used in this study.

\subsection{LSTM}
Long Short-Term Memory (LSTM) networks were introduced in 1997 by S. Hochreiter and J. Schmidhuber \cite{ref_lstm} and are an established variation of Recurrent Neural Networks (RNNs) which accounts for the RNNs' known problems regarding long-term dependency such as vanishing and exploding gradients i.e. when gradients become increasingly small or excessively high during backpropagation. 

In this work we referred to a modern version of the architecture which is well documented for both Keras \cite{keras} and PyTorch \cite{pytorch_lstm}, two of the most widely utilized machine learning libraries as of today.

\begin{figure}
    \centering
    \includegraphics[width=1\linewidth]{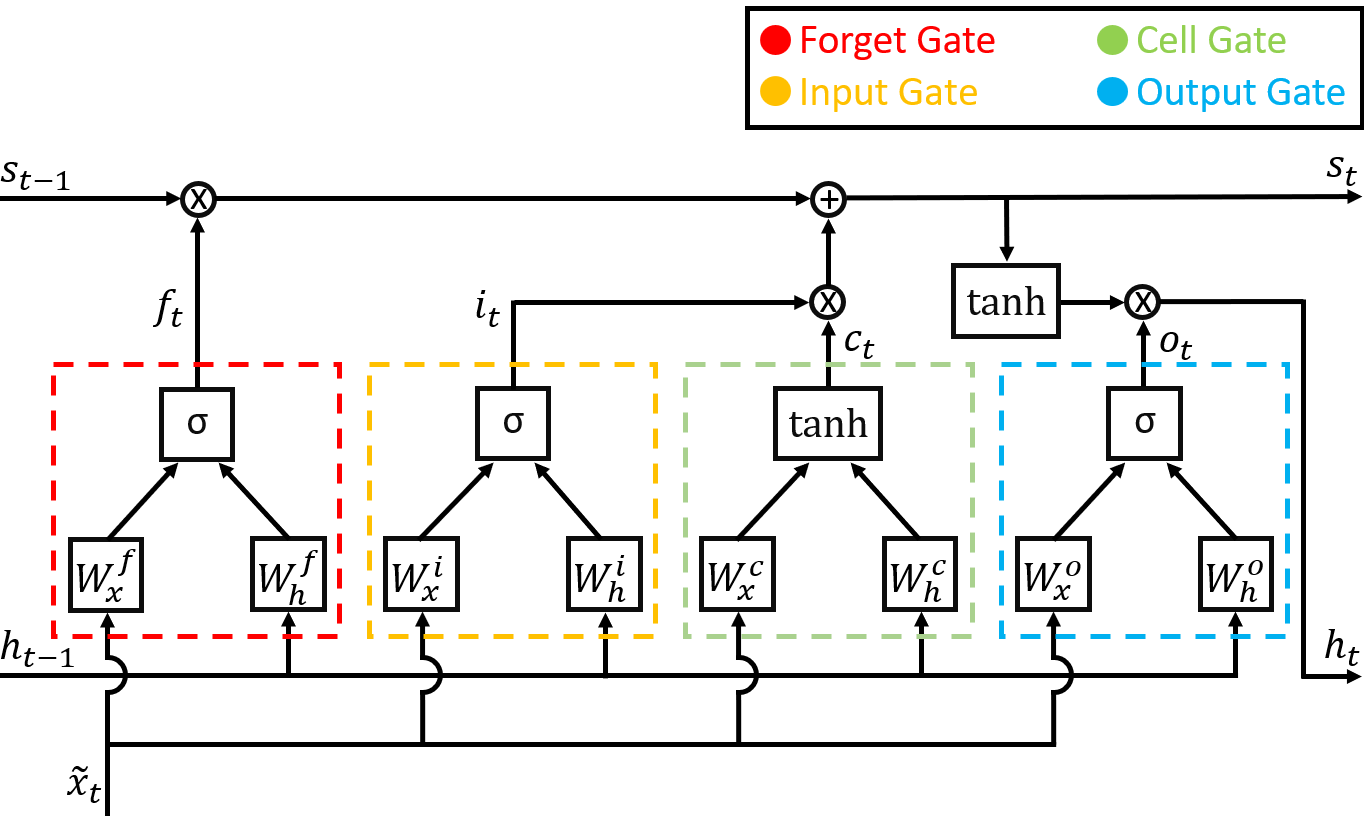}
    \caption{Structure of an LSTM's cell.}
    \label{fig:LSTM_unit}
\end{figure}

LSTMs' basic unit, which will be referred to as a "cell" from now on, is able to retain information from previous time iterations through the network (unlike the widely known neuron of feed-forward neural networks). A cell does this by operating four gates (three on some publications which do not consider the cell gate to be a gate of its own). These gates are the input gate, which controls what gets loaded into the cell, the output gate, which controls what gets passed on to the next iteration, the forget gate which controls what gets forgotten and the cell gate which "prepares" the information to be sent through the input gate. The structure of an LSTM's cell is shown in Fig. \ref{fig:LSTM_unit}.

The input to an LSTM's cell is a single feature vector (a time-step) $\tilde{x}_t\in\mathbb{R}^{d_{in}}$ from the input timeseries $\tilde{X}\in\mathbb{R}^{n \times d_{in}}$. In addition to it, at every time-step $t$ an LSTM cell also has a state $s_t$ and produces an output $h_t$, which are passed down as inputs to the next time iteration of the cell. $s_t$ and $h_t$ are the vectors containing the memory of an LSTM cell and are used to keep information stored from previous iterations of the network. The outputs of the four gates are computed as:
\begin{equation}
    f_t=\sigma (\tilde{x}_t(W_x^f)^T+h_{t-1}(W_h^f)^T+bias_f)
\end{equation}
\begin{equation}
    i_t=\sigma (\tilde{x}_t(W_x^i)^T+h_{t-1}(W_h^i)^T+bias_i)
\end{equation}
\begin{equation}
    c_t=\textnormal{tanh}(\tilde{x}_t(W_x^c)^T+h_{t-1}(W_h^c)^T+bias_c)
\end{equation}
\begin{equation}
    o_t=\sigma (\tilde{x}_t(W_x^o)^T+h_{t-1}(W_h^o)^T+bias_o)
\end{equation}
where $f_t$, $i_t$, $c_t$ and $o_t\in\mathbb{R}^{d_h}$ are respectively the outputs of the forget, input, cell and output gates and $d_h$ being the size of the LSTM layer (the structure of an LSTM layer will be discussed shortly afterwards). $W_x^f$, $W_h^f$, $W_x^i$, $W_h^i$, $W_x^c$, $W_h^c$, $W_x^o$, $W_h^o$ are weight matrices with $W_x^f$, $W_x^i$, $W_x^c$, $W_x^o\in \mathbb {R}^{d_h \times d_{in}}$ and $W_h^f$, $W_h^i$, $W_h^c$, $W_h^o\in \mathbb {R}^{d_h \times d_h}$. $bias_f$, $bias_i$, $bias_c$, $bias_o$ are learnable biases.

At each time-step the state and the output get updated as:
\begin{equation}
    s_t=s_{t-1}f_t+i_tc_t
\end{equation}
\begin{equation}
    h_t=o_t\textnormal{tanh}(s_t)
\end{equation}

\begin{figure}
    \centering
    \includegraphics[width=0.3\linewidth]{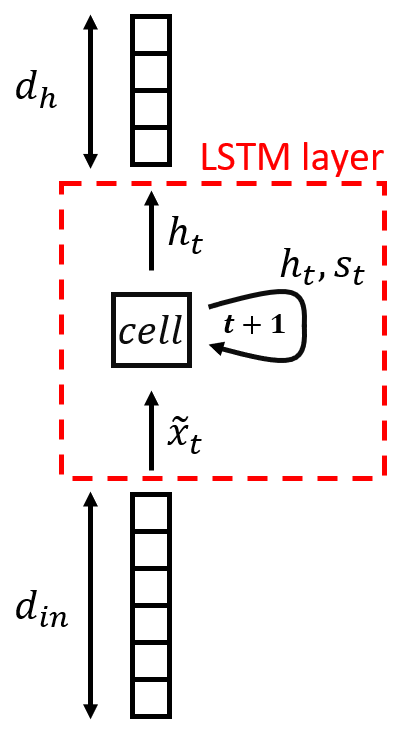}
    \caption{Structure of an LSTM's layer.}
    \label{fig:LSTM_layer_unwrap}
\end{figure}

\begin{figure}
    \centering
    \includegraphics[width=1\linewidth]{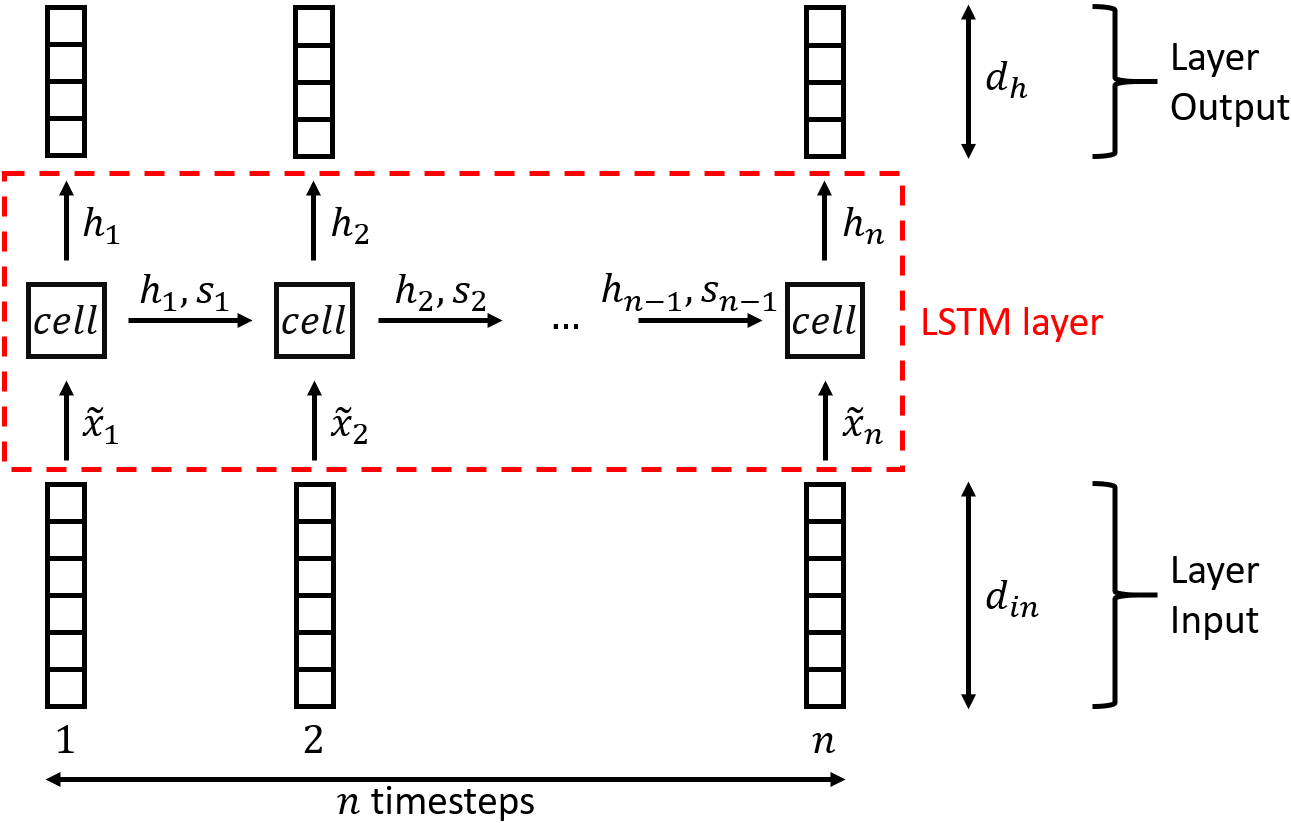}
    \caption{Structure of an LSTM's layer with the time loop unwrapped to highlight the process through the time-steps.}
    \label{fig:LSTM_layer}
\end{figure}

An LSTM layer is made up of a cell with a self-loop which gets updated at every time-step (see Fig. \ref{fig:LSTM_layer_unwrap}). At each time-step, the relative feature vector and the cell's output and state from the previous time-step are used by the cell as inputs. If the self-loop through the time-steps is unwrapped as shown in Fig. \ref{fig:LSTM_layer} it becomes clearer how an LSTM layer is essentially an LSTM cell iterated through the $n$ time-steps of an input timeseries.

The output of an LSTM layer is typically a matrix $\overline{Y}=LSTM(\overline{X})\in\mathbb {R}^{n \times d_h}$. The $i^{th}$ row of $\overline{Y}$ will be equal to $h_i$, i.e. the output of the LSTM cell at the $i^{th}$ time-step. On the other hand, if the LSTM layer is single output, $\overline{Y}=h_n$ i.e. the output of the layer is equal to the output of the LSTM cell at the last time-step.

\begin{figure}
    \centering
    \includegraphics[width=0.28\linewidth]{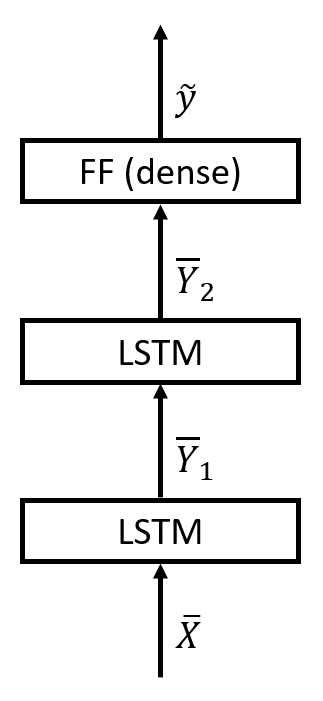}
    \caption{Structure of the LSTM network.}
    \label{fig:LSTM_network}
\end{figure}

For this paper, an LSTM network whose general structure is shown in Fig. \ref{fig:LSTM_network} was used. At first, the input passes through a number of stacked LSTM layers (for simplicity only two are shown in Fig. \ref{fig:LSTM_network}). The last of the stacked LSTM layers is single output precedes a feed-forward dense layer with a softmax activation function which produces the probabilities with which each input timeseries will lead to one of the classes (Lk, LLC or RLC) within the set maximum prediction time. Each input is assigned to the class corresponding to the highest probability. 

Three LSTM configurations were tested in this paper (LSTM 1, LSTM 2 and LSTM 3). Their structures are shown in Tab. \ref{tab:lstm}. LSTM 1 has three stacked LSTM layers, LSTM 2 and 3 only have two.

\begin{table*}
    \centering
    \begin{tabular}{cccccc}
        \multicolumn{1}{c}{}& LSTM layers & $1^{st}$ layer dimension & $2^{nd}$ layer dimension &  $3^{rd}$ layer dimension& Optimizer \\ \hline \hline
        LSTM 1 & 3 & 2 & 2 & 1 & Adam \\ \hline 
        LSTM 2 & 2 & 2 & 2 & n.a. & Adam \\ \hline 
        LSTM 3 & 2 & 2 & 1 & n.a. & Adam \\ \hline 
        \multicolumn{6}{c}{}\\
        \multicolumn{6}{c}{}\\
    \end{tabular}
    \caption{Configurations of the tested LSTMs.}
    \label{tab:lstm}
\end{table*}

\subsection{CNN}

Convolutional Neural Networks (CNN) are a family of widely used deep learning algorithms inspired by the visual perception system of living beings \cite{ref_cnn_gu.2018}. Their initial development dates back to the late 1980s \cite{ref_cnn_leCun.1989}, but CNNs gained popularity in the 1990s, particularly through their application in handwritten digit recognition \cite{ref_cnn_leCun.1998}. Thanks to their ability to capture spatial information, CNNs are especially well-suited for image classification tasks. Today, they are applied across various domains, including $\textendash$ but not limited to $\textendash$ object detection, object tracking, scene classification, and natural language processing \cite{ref_cnn_gu.2018}. CNNs were also successfully utilized for different transportation related problems including lane change intention prediction.

\begin{figure}
    \centering
    \includegraphics[width=0.99\linewidth]{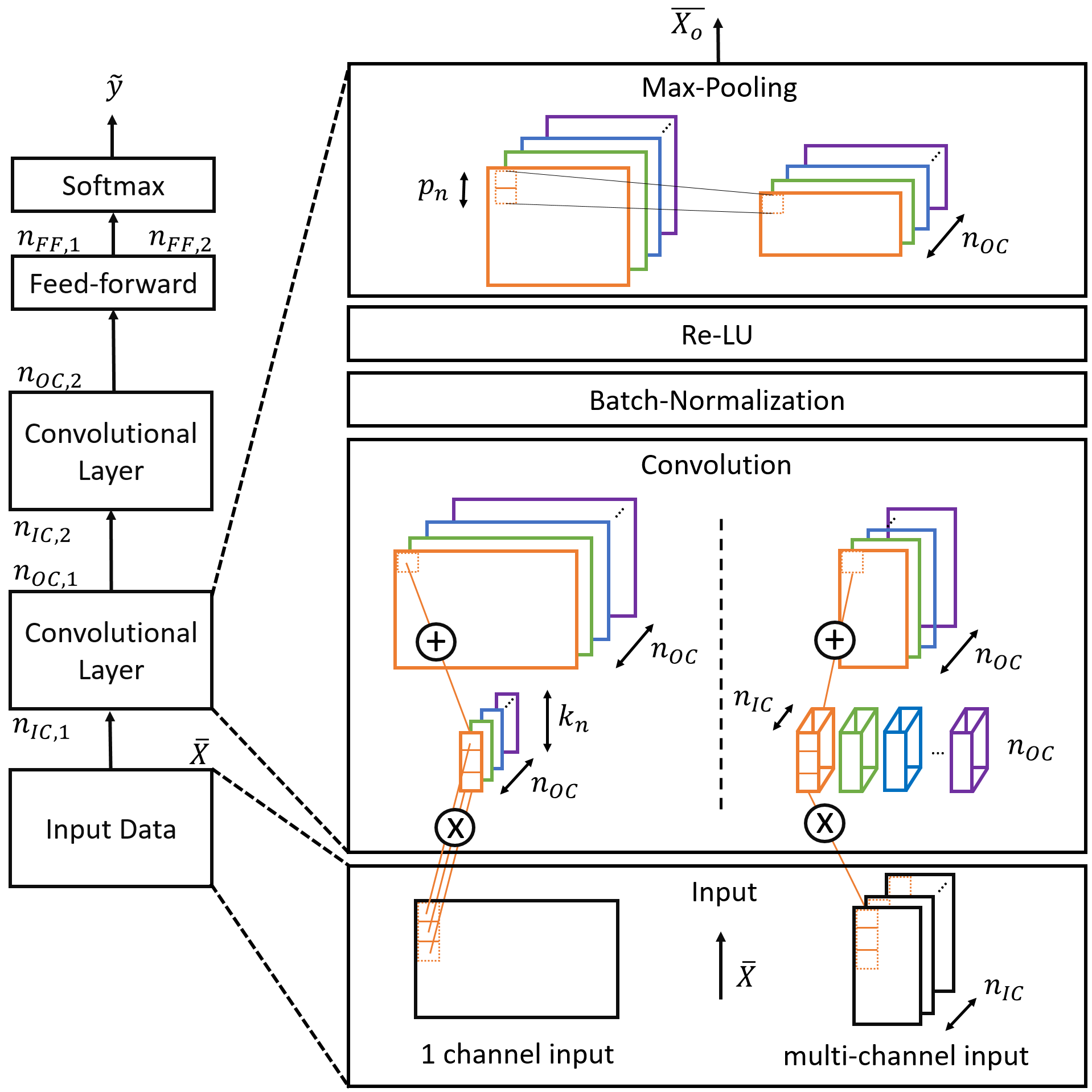}
    \caption{Structure of the CNN}
    \label{fig:cnn}
\end{figure}

The hyper-parameters of the three evaluated CNN architectures are summarized in Tab. \ref{tab:cnn}, while their structures are illustrated in Fig. \ref{fig:cnn}. In this study, the multivariate time series input data $\overline{X}$ is fed into two convolutional layers, followed by a two-layer feed-forward neural network. The fully connected layers contain $n_{FF,1}$ and $n_{FF,2}$ nodes in the first and second layer, respectively. The final output layer applies a softmax activation, yielding the vector $\tilde{y}$, which represents the predicted class probabilities for the three target classes.
Across the three tested CNN architectures, the input data is processed differently. In CNN 2 and CNN 3 the input is organized through a single channel, resulting in a tensor representation $\overline{X}_{sc} \in \mathbb{R}^{n \times d}$. In CNN 1, inspired by the structure of image data, the input $\overline{X}$ is split into multiple channels, resulting in a tensor representation $\overline{X}_{mc} \in \mathbb{R}^{n \times (d / n_{ic}) \times n_{ic}}$, where $n$ is the number of time steps, $d$ the total number of features, and $n_{ic}$ the number of input channels. This representation, $\overline{X}_{mc}$, contains separated time series data for position and speed, with one channel for the ego vehicle and one for each surrounding vehicle.

To extract temporal patterns from this input, each convolutional layer applies a set of learnable kernels. Let $\mathbf{K}^{(l)} \in \mathbb{R}^{n_{oc} \times n_{ic} \times t_l \times d_l}$ denote the set of convolutional kernels in layer $l$, where $n_{ic}$ and $n_{oc}$ are the number of input and output channels, respectively, and $t_l \times d_l$ defines the spatial dimensions of each kernel. 
Since the features capture heterogeneous aspects (e.g., positions and velocities) and lack a direct spatial correlation along the feature axis, $d_l$ is set to $1$.
Using the kernel set $\mathbf{K}^{(l)}$, each convolutional layer performs a point-wise weighted summation over the local input region, followed by batch normalization and applying a non-linearity (ReLU). Subsequently, a one-dimensional max-pooling operation is applied along the temporal axis, reducing the temporal resolution of the resulting feature map by a factor of $p_n$. This yields an output tensor $\overline{X}_{o} \in \mathbb{R}^{n_{oc} \times (t_l / p_n) \times d_l}$.

\begin{table*}
    \centering
    \begin{tabular}{cccccccccccc}
        \multicolumn{1}{c}{} & $n_{ic,1}$ & $n_{oc,1}\;(=n_{ic,2})$  & $n_{oc,2}$  & $p_n$  & $k_n$ & Batch-Norm  & $n_{FF,1}$ & $n_{FF,2}$ &  $FF_{dropout}$ & learning rate & optimizer \\ \hline \hline
        CNN 1 & 9 & 12 & 18 & 2 & 5 & True & 64 & 32 & 0.5  & 0.0001  & Adam   \\ \hline 
        CNN 2 & 1 & 12 & 18 & 2 & 3 & False & 256 & 128 & 0.5  & 0.0001  & Adam  \\ \hline 
        CNN 3 & 1 & 18 & 6 & 2 & 5 & True & 64 & 32 & 0.5  & 0.0001  & Adam   \\ \hline
        \multicolumn{7}{c}{}\\
        \multicolumn{7}{c}{}\\
    \end{tabular}
    \caption{Configurations of the tested CNNs.}
    \label{tab:cnn}
\end{table*}

\subsection{Transformer network}
Transformer Networks (TNs), sometimes simply referred to as Transformers, are a family or neural networks introduced in 2017 by Vaswani et all. \cite{ref_trans}. The key idea of TNs is to find relationships between the elements of an input sequence and exploit this information to generate an output. Typical applications of this architecture are language translation and generation but by adapting the architecture it is possible to solve classification problems as in our case.

\begin{figure}
    \centering
    \begin{subfigure}{(a)}
        \centering
        \includegraphics[height=2.3in]{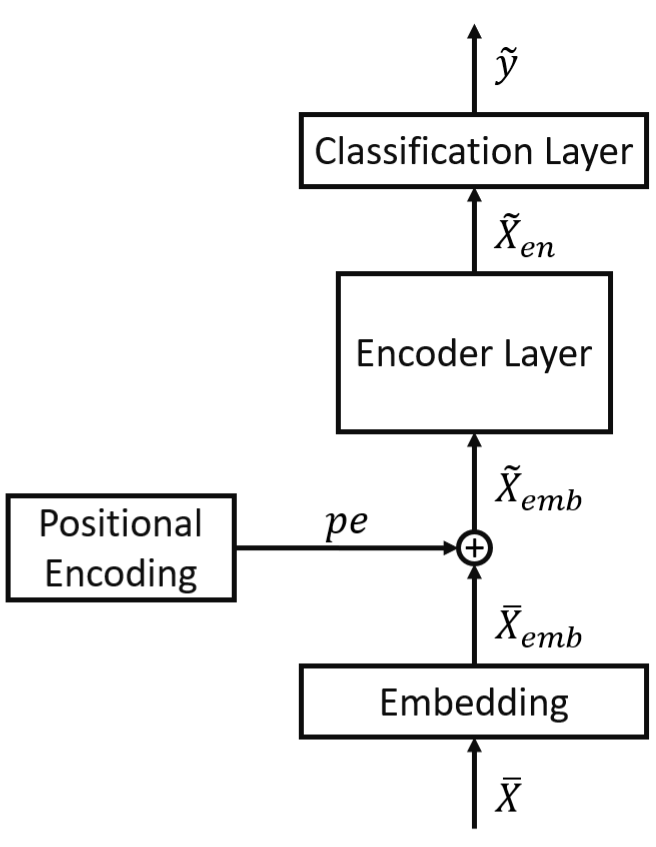}
    \end{subfigure}
    \begin{subfigure}{(b)}
        \centering
        \includegraphics[height=2.3in]{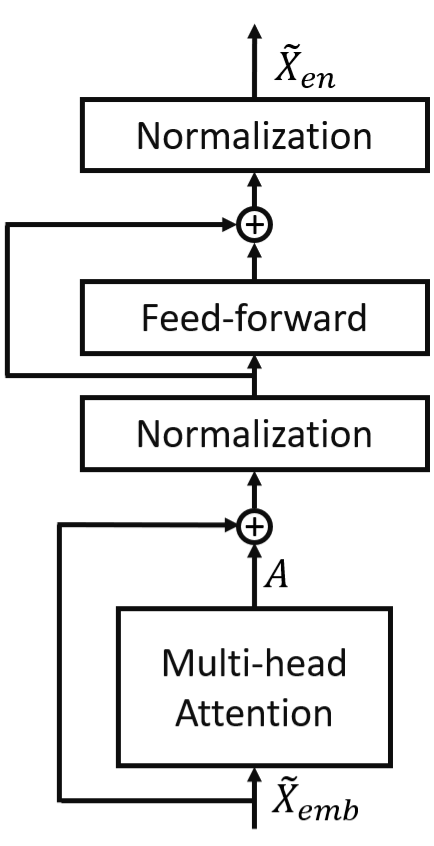}
    \end{subfigure}
    \caption{Structure of the transformer network (a) and close-up of the encoder layer (b).}
    \label{TN_structure}
\end{figure}
\begin{figure}
    \centering
    \includegraphics[width=0.4\linewidth]{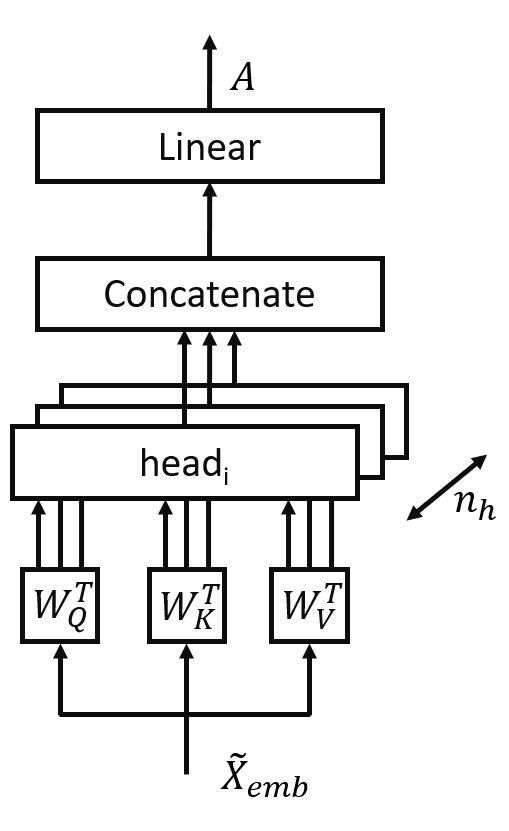}
    \caption{Structure of the multi-head attention layer.}
    \label{multi-head}
\end{figure}
TNs are typically made up of an encoder layer, which as the name suggests encodes the input data in a suitable format, and a decoder layer, which taking the output of the encoder generates an output sequence. Since in our case the problem to be solved is a classification one and there is no output sequence to be generated, no decoder is included in our architecture and a linear classification layer (which operates an affine linear transformation) is added after the encoder. The structure of our TN is shown in Fig. \ref{TN_structure}. The embedding layer consists of a linear function $f_{emb}$ that takes as an input a multivariate time-series (a trajectory segment $\overline{X}$) and reshapes each time-step to a desired dimension $d_{emb}$ (originally the dimension is the number of input features $d$) to generate an embedded input multivariate time-series $\overline{X}_{emb}$:
\begin{equation}
    \overline{X}_{emb}=f_{emb}(\overline{X})
\end{equation}
with $\overline{X} \in \mathbb{R}^{nd}$ and $\overline{X}_{emb} \in \mathbb{R}^{n \times d_{emb}}$. The positional encoding's ($pe$) purpose is to help the TN understand the structure of the input time series and it is defined as:
\begin{equation}
    pe_{i,j}=\begin{cases}
        \textnormal{sin}((i-1)/1000^{(j-1)/d_{emb}})\;\;if\;\;j\;\;is\;\;odd\\
        \textnormal{cos}((i-1)/1000^{(j-2)/d_{emb}})\;\;if\;\;j\;\;is\;\;even
    \end{cases}
\end{equation}
with $i = 1,...,n$, $j = 1,...,d_{emb}$ and $pe \in \mathbb{R}^{n \times d_{emb}}$. The positional encoding is then added (a dropout rate of $0.1$ is applied to the positional encoding) to $\overline{X}_{emb}$ and the resulting $\tilde{X}_{emb}\in\mathbb{R}^{n \times d_{emb}}$ is calculated as:
\begin{equation}
    \tilde{X}_{emb}=\overline{X}_{emb}+pe
\end{equation}
which is then fed to the encoder layer shown in Fig. \ref{TN_structure}. The first element of the encoder is the multi-head attention block, whose structure is shown is Fig. \ref{multi-head}. At first, 
$\tilde{X}_{emb}$ is projected into query, key and value matrices $Q$, $K$, $V\in\mathbb{R}^{n \times d_{emb}}$:
\begin{gather}
    Q=\tilde{X}_{emb}W_Q^T\;,\;\;W_Q\in \mathbb{R}^{d_{emb} \times d_{emb}}\\
    K=\tilde{X}_{emb}W_K^T\;,\;\;W_K\in \mathbb{R}^{d_{emb} \times d_{emb}}\\
    V=\tilde{X}_{emb}W_V^T\;,\;\;W_V\in \mathbb{R}^{d_{emb} \times d_{emb}}
\end{gather}
For each head $i=1,...,n_h$ of the multi-head attention block the relative attention $a_i$ is computed as:
\begin{gather}
    a_i=softmax(\frac{QW_{q,i}^T(KW_{k,i}^T)^T}{\sqrt{d_h}})VW_{v,i}^T
\end{gather}
where $W_{q,i}$, $W_{k,i}$, $W_{v,i}\in\mathbb{R}^{d_h \times d_{emb}}$ and $d_h=\lfloor d_{emb}/n_h\rfloor$ for $i=1,...,(n_h-1)$. For $i=n_h$ instead, $W_{q,i}$, $W_{k,i}$, $W_{v,i}\in\mathbb{R}^{d_r \times d_{emb}}$ and $d_r=d_{emb}-n_hd_h$. The resulting attentions are then concatenated and linearly combined to generate the multi-head attention $A$:
\begin{equation}
    A=[a_1 ... a_{n_h}]W_A^T
\end{equation}
where $W_A\in\mathbb{R}^{d_{emb} \times d_{emb}}$ and $A\in\mathbb{R}^{n \times d_{emb}}$. The output $\overline{X}_{en}\in\mathbb{R}^{n \times d_{emb}}$ of the encoder layer is then computed as:
\begin{equation}
    \tilde{X}_{en}=Norm(Norm(A+\tilde{X}_{emb})+FF(Norm(A+\tilde{X}_{emb})))
\end{equation}
where $Norm()$ indicates a normalization layer and $FF()$ indicates a feed-forward layer of width $w_{FF}$.

Finally, $\tilde{X}_{en}$ passes through a linear classification layer which outputs $\tilde y$ which is a vector containing three values, one per class. Each sample is assigned to the class for which the respective output value is the highest among the three output values.

Three configurations of the described transformers were tested with varying design parameters as described in Tab. \ref{tab:transformers}. The weight decay was adjusted to avoid over-fitting and an appropriate learning rate was selected to avoid instability during the training process. TN 1 is designed to be as simple as possible without losing acceptable performances while the complexity of the network was augmented increasingly in TN 2 and TN 3 to observe which effects the design parameters would've had on the results. Previously, when describing the structure of a transformer network, the number of encoder layers was assumed to be only one out of simplicity; when more encoder layers are present, they are simply stacked one after each other.

\begin{table*}
    \centering
    \begin{tabular}{cccccccc}
        
        \multicolumn{1}{c}{}& Encoder layers & $n_h$ & $d_{emb}$ & $w_{FF}$ & Weight decay & Learning rate & Optimizer\\ \hline \hline
        TN 1 & 1 & 16 & 16 & 16 & 0.004 & 0.0007 & Adam\\ \hline 
        TN 2 & 1 & 16 & 128 & 64 & 0.004 & 0.0007 & Adam \\ \hline 
        TN 3 & 4 & 16 & 128 & 64 & 0.004 & 0.0007 & Adam \\ \hline 
        \multicolumn{7}{c}{}\\
        \multicolumn{7}{c}{}\\
    \end{tabular}
    \caption{Configurations of the tested transformers.}
    \label{tab:transformers}
\end{table*}

\section{Results}\label{sec:results}
The results of our tests will be presented in this section. At first, the metrics used in the discussion of the results will be introduced. After that, the results for the tested LSTMs, CNNs and TNs will be presented. Finally, the best LSTM, the best CNN and the best TN will be compared to each other.

\subsection{Metrics}
Before introducing the metrics, a brief explanation of what a confusion matrix is will be provided here for clarity. A confusion matrix, as shown in Tab. \ref{tab:confusion_structure}, is a matrix in which the results of a classification problem over the test dataset are recorded. In Tab. \ref{tab:confusion_structure} each row corresponds to the true classes assigned to the samples, while columns correspond to the predicted classes. In this case, as already stated earlier, the classes are LK, LLC and RLC. Keeping this in mind, values "A", "E" and "I" represent the samples which were correctly predicted while for example value "D" represents the number of LLC samples which got misclassified by the tested algorithm as LKs. Symbols like "A", "E", "D" etc. will be used in the formulas of the metrics with reference to the symbols used in Tab. \ref{tab:confusion_structure} if not otherwise specified. The metrics used in this paper can now be introduced:\\

\begin{enumerate}
    \item \textbf{Accuracy}\\ Accuracy (Acc) shows the percentage of correctly classified samples over the total number of samples in the test dataset: \begin{equation}
        Acc = 100\frac{Correctly\;\;classified\;\;samples}{Total\;\;number\;\;of\;\;samples}
    \end{equation}
    
    \item \textbf{$\Delta_{acc}$ score}\\ $\Delta_{acc}$ score is defined as the difference between the accuracies calculated on the train dataset (computed in the same way as they are for the test dataset) and the test dataset. It is used to show the overfitting of a specific method i.e. the tendency of a method to fit excessively the train data having good results on it but having bad results on the test data. The greater the $\Delta_{acc}$ score, the greater the overfitting: \begin{equation}
        \Delta_{acc} = Acc_{train} - Acc_{test}
    \end{equation}
    
    \item \textbf{Recall}\\Recall (Re) indicates, for each "true" class, the percentage of samples for that specific class which got correctly classified:
    \begin{gather}
        Re_{LK}=100\frac{A}{A+B+C}\\
        Re_{LLC}=100\frac{E}{D+E+F}\\
        Re_{RLC}=100\frac{I}{G+H+I}
    \end{gather}
    
    \item \textbf{Precision}\\Precision (Pr) indicates, for each "predicted" class, the percentage of samples for that specific class which got correctly classified:
    \begin{gather}
        Pr_{LK}=100\frac{A}{A+D+G}\\
        Pr_{LLC}=100\frac{E}{B+E+H}\\
        Pr_{RLC}=100\frac{I}{C+F+I}
    \end{gather}

    \item \textbf{$F_1$ score}\\The $F_1$ score ($F_1$) represents, for each class, the harmonic mean between precision and recall for said class:
    \begin{gather}
        F_{1,LK}=\frac{2Pr_{LK}Re_{LK}}{Pr_{LK}+Re_{LK}}\\
        F_{1,LLC}=\frac{2Pr_{LLC}Re_{LLC}}{Pr_{LLC}+Re_{LLC}}\\
        F_{1,RLC}=\frac{2Pr_{RLC}Re_{RLC}}{Pr_{RLC}+Re_{RLC}}
    \end{gather}
    
\end{enumerate}

Recall and precision were not directly employed in the evaluation of the results but still had to be introduced because they are used to calculate $F_1$ scores.

\begin{table}[]
    \centering
    \begin{tabular}{ccc|c|c|c}
        \multicolumn{3}{c}{}&\multicolumn{3}{c}{Predicted} \\ \\
        &&& LK & LLC & RLC  \\ \cline{3-6}
        \multirow{3}{*}{\rotatebox{90}{True}}&&LK & A & B & C \\ \cline{3-6}
        &&LLC & D & E & F \\ \cline{3-6}
        &&RLC & G & H & I \\ \cline{3-6}
    \end{tabular}
    \caption{Structure of the confusion matrices which were used to evaluate the results. "A", "E" and "I" represent respectively the correctly identified LK, LLC and RLC samples in the test dataset.}
    \label{tab:confusion_structure}
\end{table}

\subsection{LSTM results}
The three proposed LSTM architectures (see Tab. \ref{tab:lstm}) were tested on the test dataset (see Chapter \ref{sec:data}) for varying values of $\Delta t_{p,\textnormal{MAX}}$ and $\Delta t_o$. The accuracies and $F_1$ scores are shown in Tab. \ref{tab:lstm_results}. A deterioration of the results can be observed for all three configurations and for all values of the observation window when increasing the maximum prediction time. For LSTM 1, the most complex among the three configurations, a longer observation window translates to better accuracy with the exception of the case $\Delta t_{p,\textnormal{MAX}}=5\textnormal{s}$ in which a shorter observation window seems to achieve better performances, especially for the prediction of LKs and LLCs (as indicated by the higher $F_1$ scores for these two classes). For LSTM 2 and LSTM 3 instead, increasing the observation window does not seem to improve the result significantly if at all. 

Overall, LSTM 2 has on average better accuracy and $F_1$ score values with respect to the other two configurations. This is especially interesting considering that it has only two LSTM layers compared to the three layers of LSTM 1.

\begin{table*}
    \centering
    \begin{tabular}{c c|c|c c c|c c c|c c c|}
        \multicolumn{3}{c}{}& \multicolumn{9}{c}{Observation Window} \\ 
        
        \multicolumn{3}{c}{}& \multicolumn{3}{c}{$\Delta t_o$: $1\textnormal{s}$} & \multicolumn{3}{c}{$\Delta t_o$: $2\textnormal{s}$} & \multicolumn{3}{c}{$\Delta t_o$: $3\textnormal{s}$}\\ \cline{4-12}

        \multicolumn{3}{c|}{}&&&&&&&&&\\
        
        \multicolumn{3}{c|}{}& LSTM 1 & LSTM 2 & LSTM 3 & LSTM 1 & LSTM 2 & LSTM 3 & LSTM 1 & LSTM 2 & LSTM 3  \\ \cline{3-12}
        
        \multirow{16}{*}{\rotatebox{90}{Maximum Prediction Time}}&\multirow{4}{*}{$\Delta t_{p,\textnormal{MAX}}$: $3\textnormal{s}$} & 
        Acc & $91.38\%$ & $\boldsymbol{94.09\%}$ & $93.80\%$ & $94.56\%$ & $94.35\%$ & $\boldsymbol{95.69\%}$ & $94.25\%$ & $\boldsymbol{95.22\%}$ & $93.01\%$ \\
        && $F_{1,LK}$ & $92.86\%$ & $\boldsymbol{94.29\%}$ & $93.95\%$ & $94.57\%$ & $94.37\%$ & $\boldsymbol{95.64\%}$ & $94.28\%$ & $\boldsymbol{95.23\%}$ & $93.42\%$ \\
        && $F_{1,LLC}$ & $85.71\%$ & $92.28\%$ & $\boldsymbol{93.12\%}$ & $95.65\%$ & $93.89\%$ & $\boldsymbol{95.97\%}$ & $93.94\%$ & $\boldsymbol{95.43\%}$ & $94.20\%$ \\
        && $F_{1,RLC}$ & $93.44\%$ & $\boldsymbol{95.17\%}$ & $94.07\%$ & $94.47\%$ & $94.70\%$ & $\boldsymbol{95.55\%}$ & $94.46\%$ & $\boldsymbol{95.01\%}$ & $91.36\%$ \\\cline{3-12}
        
        &\multirow{4}{*}{$\Delta t_{p,\textnormal{MAX}}$: $4\textnormal{s}$} & 
        Acc & $88.59\%$ & $\boldsymbol{89.84\%}$ & $88.32\%$ & $89.23\%$ & $\boldsymbol{89.64\%}$ & $88.36\%$ & $89.39\%$ & $89.43\%$ & $\boldsymbol{89.71\%}$ \\
        && $F_{1,LK}$ & $88.75\%$ & $\boldsymbol{89.99\%}$ & $88.43\%$ & $89.47\%$ & $\boldsymbol{90.00\%}$ & $88.65\%$ & $89.87\%$ & $89.81\%$ & $\boldsymbol{90.10\%}$ \\
        && $F_{1,LLC}$ & $88.57\%$ & $\boldsymbol{90.87\%}$ & $88.28\%$ & $89.75\%$ & $\boldsymbol{89.79\%}$ & $87.89\%$ & $89.52\%$ & $89.84\%$ & $\boldsymbol{89.94\%}$ \\
        && $F_{1,RLC}$ & $88.31\%$ & $\boldsymbol{88.76\%}$ & $88.16\%$ & $88.33\%$ & $\boldsymbol{88.82\%}$ & $88.21\%$ & $88.28\%$ & $88.31\%$ & $\boldsymbol{88.75\%}$ \\\cline{3-12}
        
        &\multirow{4}{*}{$\Delta t_{p,\textnormal{MAX}}$: $5\textnormal{s}$} & 
        Acc & $\boldsymbol{85.34\%}$ & $83.75\%$ & $83.86\%$ & $80.61\%$ & $80.98\%$ & $\boldsymbol{82.25\%}$ & $\boldsymbol{83.36\%}$ & $81.94\%$ & $83.05\%$ \\
        && $F_{1,LK}$ & $\boldsymbol{86.09\%}$ & $84.85\%$ & $84.71\%$ & $81.57\%$ & $82.33\%$ & $\boldsymbol{83.36\%}$ & $\boldsymbol{84.27\%}$ & $83.22\%$ & $83.97\%$ \\
        && $F_{1,LLC}$ & $\boldsymbol{87.27\%}$ & $84.09\%$ & $85.87\%$ & $78.66\%$ & $78.17\%$ & $\boldsymbol{78.95\%}$ & $82.37\%$ & $79.33\%$ & $\boldsymbol{82.9\%}$ \\
        && $F_{1,RLC}$ & $\boldsymbol{82.23\%}$ & $81.27\%$ & $80.47\%$ & $80.35\%$ & $80.57\%$ & $\boldsymbol{82.69\%}$ & $\boldsymbol{82.32\%}$ & $81.60\%$ & $81.10\%$ \\ \cline{3-12}
        
        &\multirow{4}{*}{$\Delta t_{p,\textnormal{MAX}}$: $6\textnormal{s}$} & 
        Acc & $75.08\%$ & $\boldsymbol{81.35\%}$ & $76.93\%$ & $78.44\%$ & $78.54\%$ & $\boldsymbol{79.50\%}$ & $80.12\%$ & $\boldsymbol{80.77\%}$ & $76.08\%$ \\
        && $F_{1,LK}$ & $80.65\%$ & $\boldsymbol{82.29\%}$ & $79.25\%$ & $80.52\%$ & $80.43\%$ & $\boldsymbol{81.19\%}$ & $81.08\%$ & $\boldsymbol{81.90\%}$ & $77.67\%$ \\
        && $F_{1,LLC}$ & $76.38\%$ & $\boldsymbol{83.41\%}$ & $72.98\%$ & $75.99\%$ & $78.67\%$ & $\boldsymbol{81.92\%}$ & $78.25\%$ & $\boldsymbol{81.72\%}$ & $72.75\%$ \\
        && $F_{1,RLC}$ & $62.60\%$ & $\boldsymbol{77.79\%}$ & $75.12\%$ & $\boldsymbol{75.85\%}$ & $74.18\%$ & $73.33\%$ & $\boldsymbol{79.89\%}$ & $77.42\%$ & $75.90\%$ \\ \cline{3-12}
        
        \multicolumn{7}{c}{}\\
        \multicolumn{7}{c}{}\\
    \end{tabular}
    \caption{Accuracy (Acc) and F1 scores of the tested LSTMs in relation to the observation window and the maximum prediction time.}
    \label{tab:lstm_results}
\end{table*}

\subsection{CNN results}
The three proposed CNN architectures (see Tab. \ref{tab:cnn}) were tested on the test dataset (see Chapter \ref{sec:data}) for varying values of $\Delta t_{p,\textnormal{MAX}}$ and $\Delta t_o$. The accuracies and $F_1$ scores are shown in Tab. \ref{tab:cnn_results}. A deterioration of the results can be observed for all three configurations and for all values of the observation window when increasing the maximum prediction time. For all three configurations, increasing the observation window did not appear to bring performance improvements. On the other hand, for CNN 3 it appears that smaller observation windows led to slightly better results consistently through the different values of maximum prediction time. For higher maximum prediction times all of the tested CNNs had worse performances when prediction RLCs with respect to LKs and LLCs (lower $F_{1,RLC}$ scores) for any value of the observation window, but it seems that this effect was particularly relevant for lower observation windows.

Overall, CNN 3 resulted in higher accuracies and $F_1$ scores, making it the best performing CNN among the tested ones.
Overall
\begin{table*}
    \centering
    \begin{tabular}{c c|c|c c c|c c c|c c c|}
        \multicolumn{3}{c}{}& \multicolumn{9}{c}{Observation Window} \\
        
        \multicolumn{3}{c}{}& \multicolumn{3}{c}{$\Delta t_o$: $1\textnormal{s}$} & \multicolumn{3}{c}{$\Delta t_o$: $2\textnormal{s}$} & \multicolumn{3}{c}{$\Delta t_o$: $3\textnormal{s}$}\\ \cline{4-12}
        
        \multicolumn{3}{c|}{}&&&&&&&&&\\
        
        \multicolumn{3}{c|}{}& $\:$CNN 1$\;$ & $\;$CNN 2$\;$ & $\;$CNN 3$\:$ & $\:$CNN 1$\;$ & $\;$CNN 2$\;$ & $\;$CNN 3$\:$ & $\:$CNN 1$\;$ & $\;$CNN 2$\;$ & $\;$CNN 3$\:$  \\ \cline{3-12}
        
        \multirow{16}{*}{\rotatebox{90}{Maximum Prediction Time}}&\multirow{4}{*}{$\Delta t_{p,\textnormal{MAX}}$: $3\textnormal{s}$} & 
        Acc & $94.96\%$ & $95.43\%$ & $\boldsymbol{97.50\%}$ & $95.16\%$ & $94.85\%$ & $\boldsymbol{96.99\%}$ & $94.39\%$ & $94.49\%$ & $\boldsymbol{96.97\%}$\\
        && $F_{1,LK}$ & $96.08\%$ & $95.62\%$ & $\boldsymbol{97.57\%}$ & $95.14\%$ & $95.63\%$ & $\boldsymbol{96.96\%}$ & $94.40\%$ & $94.50\%$ & $\boldsymbol{96.94\%}$ \\
        && $F_{1,LLC}$ & $93.56\%$ & $95.76\%$ & $\boldsymbol{97.28\%}$ & $95.15\%$ & $96.26\%$ & $\boldsymbol{97.21\%}$ & $95.97\%$ & $94.41\%$ & $\boldsymbol{97.26\%}$ \\
        && $F_{1,RLC}$ & $95.86\%$ & $94.81\%$ & $\boldsymbol{97.57\%}$ & $95.20\%$ & $95.24\%$ & $\boldsymbol{96.90\%}$ & $93.05\%$ & $94.54\%$ & $\boldsymbol{96.78\%}$ \\\cline{3-12}
        
        &\multirow{4}{*}{$\Delta t_{p,\textnormal{MAX}}$: $4\textnormal{s}$} & 
        Acc & $88.92\%$ & $90.15\%$ & $\boldsymbol{91.53\%}$ & $87.29\%$ & $90.40\%$ & $\boldsymbol{91.19\%}$ & $86.89\%$ & $88.73\%$ & $\boldsymbol{91.55\%}$ \\
        && $F_{1,LK}$ & $89.03\%$ & $90.14\%$ & $\boldsymbol{91.65\%}$ & $87.74\%$ & $90.68\%$ & $\boldsymbol{91.45\%}$ & $87.57\%$ & $89.10\%$ & $\boldsymbol{91.86\%}$ \\
        && $F_{1,LLC}$ & $88.66\%$ & $90.07\%$ & $\boldsymbol{91.22\%}$ & $85.23\%$ & $89.65\%$ & $\boldsymbol{89.81\%}$ & $84.83\%$ & $88.93\%$ & $\boldsymbol{90.62\%}$ \\
        && $F_{1,RLC}$ & $88.91\%$ & $89.75\%$ & $\boldsymbol{91.56\%}$ & $88.17\%$ & $90.46\%$ & $\boldsymbol{91.81\%}$ & $87.23\%$ & $87.85\%$ & $\boldsymbol{91.75\%}$ \\ \cline{3-12}
        
        &\multirow{4}{*}{$\Delta t_{p,\textnormal{MAX}}$: $5\textnormal{s}$} & 
        Acc & $82.34\%$ & $85.23\%$ & $\boldsymbol{86.26\%}$ & $82.17\%$ & $85.45\%$ & $\boldsymbol{86.02\%}$ & $\boldsymbol{85.60\%}$ & $85.24\%$ & $85.08\%$ \\
        && $F_{1,LK}$ & $83.13\%$ & $85.46\%$ & $\boldsymbol{87.04\%}$ & $83.47\%$ & $86.23\%$ & $\boldsymbol{86.75\%}$ & $84.02\%$ & $\boldsymbol{85.69\%}$ & $84.62\%$ \\
        && $F_{1,LLC}$ & $82.00\%$ & $\boldsymbol{86.49\%}$ & $83.98\%$ & $80.78\%$ & $86.73\%$ & $\boldsymbol{87.26\%}$ & $79.65\%$ & $\boldsymbol{85.77\%}$ & $82.93\%$ \\ 
        && $F_{1,RLC}$ & $81.01\%$ & $83.86\%$ & $\boldsymbol{86.60\%}$ & $80.78\%$ & $82.85\%$ & $\boldsymbol{83.60\%}$ & $82.32\%$ & $\boldsymbol{83.85\%}$ & $83.40\%$ \\ \cline{3-12}
        
        &\multirow{4}{*}{$\Delta t_{p,\textnormal{MAX}}$: $6\textnormal{s}$} & 
        Acc & $77.66\%$ & $\boldsymbol{82.62\%}$ & $82.54\%$ & $76.1\%$ & $\boldsymbol{81.73\%}$ & $81.43\%$ & $77.38\%$ & $\boldsymbol{83.05\%}$ & $80.57\%$ \\
        && $F_{1,LK}$ & $78.79\%$ & $\boldsymbol{83.73\%}$ & $83.66\%$ & $78.20\%$ & $82.80\%$ & $\boldsymbol{83.09\%}$ & $78.60\%$ & $\boldsymbol{83.91\%}$ & $82.78\%$ \\
        && $F_{1,LLC}$ & $78.44\%$ & $\boldsymbol{83.18\%}$ & $81.51\%$ & $74.81\%$ & $\boldsymbol{83.19\%}$ & $81.83\%$ & $75.20\%$ & $\boldsymbol{82.89\%}$ & $81.97\%$ \\ 
        && $F_{1,RLC}$ & $74.92\%$ & $80.15\%$ & $\boldsymbol{81.12\%}$ & $72.97\%$ & $\boldsymbol{78.11\%}$ & $77.30\%$ & $76.99\%$ & $81.43\%$ & $\boldsymbol{81.55\%}$ \\ \cline{3-12}
        
        \multicolumn{7}{c}{}\\
        \multicolumn{7}{c}{}\\
    \end{tabular}
    \caption{Accuracy (Acc) and F1 scores of the tested CNNs in relation to the observation window and the maximum prediction time.}
    \label{tab:cnn_results}
\end{table*}

\subsection{TN results}
The three proposed TN architectures (see Tab. \ref{tab:transformers}) were tested on the test dataset (see Chapter \ref{sec:data}) for varying values of $\Delta t_{p,\textnormal{MAX}}$ and $\Delta t_o$. The accuracies and $F_1$ scores are shown in Tab. \ref{tab:trans_results}. A deterioration of the results can be observed for all three configurations and for all values of the observation window when increasing the maximum prediction time. While the length of the observation window did not seem to influence the results at lower maximum prediction windows, for $\Delta t_{p,\textnormal{MAX}}=6\textnormal{s}$ a smaller observation seems to lead to higher performances. As for CNNs and LSTMs, for higher maximum prediction times RLCs seem to be harder to predict with respect to LKs and LLCs for all the tested TNs (e.g. for $\Delta t_{p,\textnormal{MAX}}=6\textnormal{s}$, $F_1$ scores for RLCs are clearly lower in most cases than those of LLCs and LKs). 

Overall, TN 2 has on average better accuracy and $F_1$ score values with respect to the other two configurations, which could be expected for TN 1, it being extremely simple, but which was surprising for TN 3 which has three more encoder layers with respect to TN 2.

\begin{table*}
    \centering
    \begin{tabular}{c c|c|c c c|c c c|c c c|}
        \multicolumn{3}{c}{}& \multicolumn{9}{c}{Observation Window} \\
        
        \multicolumn{3}{c}{}& \multicolumn{3}{c}{$\Delta t_o$: $1\textnormal{s}$} & \multicolumn{3}{c}{$\Delta t_o$: $2\textnormal{s}$} & \multicolumn{3}{c}{$\Delta t_o$: $3\textnormal{s}$}\\ \cline{4-12}
        
        \multicolumn{3}{c|}{}&&&&&&&&&\\
        
        \multicolumn{3}{c|}{}& $\;\;$TN 1$\:\:\:$ & $\:\:\:$TN 2$\:\:\:$ & $\:\:\:$TN 3$\;\;$ & $\;\;$TN 1$\:\:\:$ & $\:\:\:$TN 2$\:\:\:$ & $\:\:\:$TN 3$\;\;$ & $\;\;$TN 1$\:\:\:$ & $\:\:\:$TN 2$\:\:\:$ & $\:\:\:$TN 3$\;\;$  \\ \cline{3-12}
        
        \multirow{16}{*}{\rotatebox{90}{Maximum Prediction Time}}&\multirow{4}{*}{$\Delta t_{p,\textnormal{MAX}}$: $3\textnormal{s}$} & 
        Acc & $\boldsymbol{96.35\%}$ & $96.24\%$ & $96.03\%$ & $95.99\%$ & $96.70\%$ & $\boldsymbol{96.91\%}$ & $95.73\%$ & $\boldsymbol{96.73\%}$ & $96.39\%$ \\
        && $F_{1,LK}$ & $\boldsymbol{96.47\%}$ & $96.35\%$ & $96.06\%$ & $96.00\%$ & $96.66\%$ & $\boldsymbol{96.86\%}$ & $95.73\%$ & $\boldsymbol{96.80\%}$ & $96.34\%$ \\
        && $F_{1,LLC}$ & $\boldsymbol{95.74\%}$ & $95.70\%$ & $94.66\%$ & $96.28\%$ & $97.00\%$ & $\boldsymbol{97.18\%}$ & $95.73\%$ & $96.65\%$ & $\boldsymbol{96.95\%}$ \\
        && $F_{1,RLC}$ & $96.61\%$ & $96.48\%$ & $\boldsymbol{97.13\%}$ & $95.74\%$ & $96.53\%$ & $\boldsymbol{96.78\%}$ & $95.74\%$ & $\boldsymbol{96.80\%}$ & $95.96\%$ \\ \cline{3-12}
        
        &\multirow{4}{*}{$\Delta t_{p,\textnormal{MAX}}$: $4\textnormal{s}$} & 
        Acc & $89.63\%$ & $\boldsymbol{91.26\%}$ & $91.03\%$ & $90.77\%$ & $\boldsymbol{92.53\%}$ & $91.26\%$ & $90.57\%$ & $\boldsymbol{92.70\%}$ & $92.50\%$ \\
        && $F_{1,LK}$ & $90.10\%$ & $\boldsymbol{91.27\%}$ & $90.99\%$ & $91.01\%$ & $\boldsymbol{92.66\%}$ & $91.13\%$ & $91.00\%$ & $\boldsymbol{92.93\%}$ & $92.75\%$ \\
        && $F_{1,LLC}$ & $88.31\%$ & $\boldsymbol{91.64\%}$ & $91.39\%$ & $89.10\%$ & $\boldsymbol{92.84\%}$ & $90.99\%$ & $89.25\%$ & $\boldsymbol{93.28\%}$ & $92.63\%$ \\
        && $F_{1,RLC}$ & $89.75\%$ & $\boldsymbol{90.96\%}$ & $90.80\%$ & $91.66\%$ & $\boldsymbol{92.03\%}$ & $91.72\%$ & $90.84\%$ & $91.76\%$ & $\boldsymbol{91.90\%}$ \\ \cline{3-12}
        
        &\multirow{4}{*}{$\Delta t_{p,\textnormal{MAX}}$: $5\textnormal{s}$} & 
        Acc & $84.54\%$ & $87.47\%$ & $\boldsymbol{87.78\%}$ & $85.86\%$ & $\boldsymbol{87.75\%}$ & $86.23\%$ & $83.00\%$ & $\boldsymbol{87.37\%}$ & $85.29\%$ \\
        && $F_{1,LK}$ & $85.50\%$ & $87.66\%$ & $\boldsymbol{88.27\%}$ & $86.88\%$ & $\boldsymbol{88.37\%}$ & $86.67\%$ & $84.38\%$ & $\boldsymbol{88.12\%}$ & $85.69\%$ \\
        && $F_{1,LLC}$ & $82.21\%$ & $\boldsymbol{89.22\%}$ & $88.96\%$ & $87.48\%$ & $\boldsymbol{89.17\%}$ & $87.91\%$ & $79.39\%$ & $\boldsymbol{87.74\%}$ & $86.01\%$ \\
        && $F_{1,RLC}$ & $84.50\%$ & $85.66\%$ & $\boldsymbol{85.81\%}$ & $82.25\%$ & $\boldsymbol{85.36\%}$ & $84.07\%$ & $83.21\%$ & $\boldsymbol{85.42\%}$ & $83.81\%$ \\ \cline{3-12}
        
        &\multirow{4}{*}{$\Delta t_{p,\textnormal{MAX}}$: $6\textnormal{s}$} & 
        Acc & $82.75\%$ & $\boldsymbol{85.16\%}$ & $84.30\%$ & $79.96\%$ & $\boldsymbol{83.92\%}$ & $83.41\%$ & $81.16\%$ & $82.79\%$ & $\boldsymbol{82.99\%}$ \\
        && $F_{1,LK}$ & $84.06\%$ & $\boldsymbol{86.19\%}$ & $85.05\%$ & $81.79\%$ & $84.77\%$ & $\boldsymbol{84.87\%}$ & $82.51\%$ & $\boldsymbol{83.82\%}$ & $83.24\%$ \\
        && $F_{1,LLC}$ & $84.77\%$ & $\boldsymbol{86.07\%}$ & $84.61\%$ & $77.62\%$ & $\boldsymbol{85.87\%}$ & $84.02\%$ & $79.20\%$ & $82.42\%$ & $\boldsymbol{85.42\%}$ \\
        && $F_{1,RLC}$ & $78.20\%$ & $82.33\%$ & $\boldsymbol{82.66\%}$ & $78.12\%$ & $\boldsymbol{80.43\%}$ & $79.62\%$ & $80.05\%$ & $\boldsymbol{81.09\%}$ & $80.10\%$ \\\cline{3-12}
        
        \multicolumn{7}{c}{}\\
        \multicolumn{7}{c}{}\\
    \end{tabular}
    \caption{Accuracy (Acc) and F1 scores of the tested transformers in relation to the observation window and the maximum prediction time.}
    \label{tab:trans_results}
\end{table*}

\subsection{Final Comparison}
\begin{figure}
    \centering
    \includegraphics[width=1\linewidth]{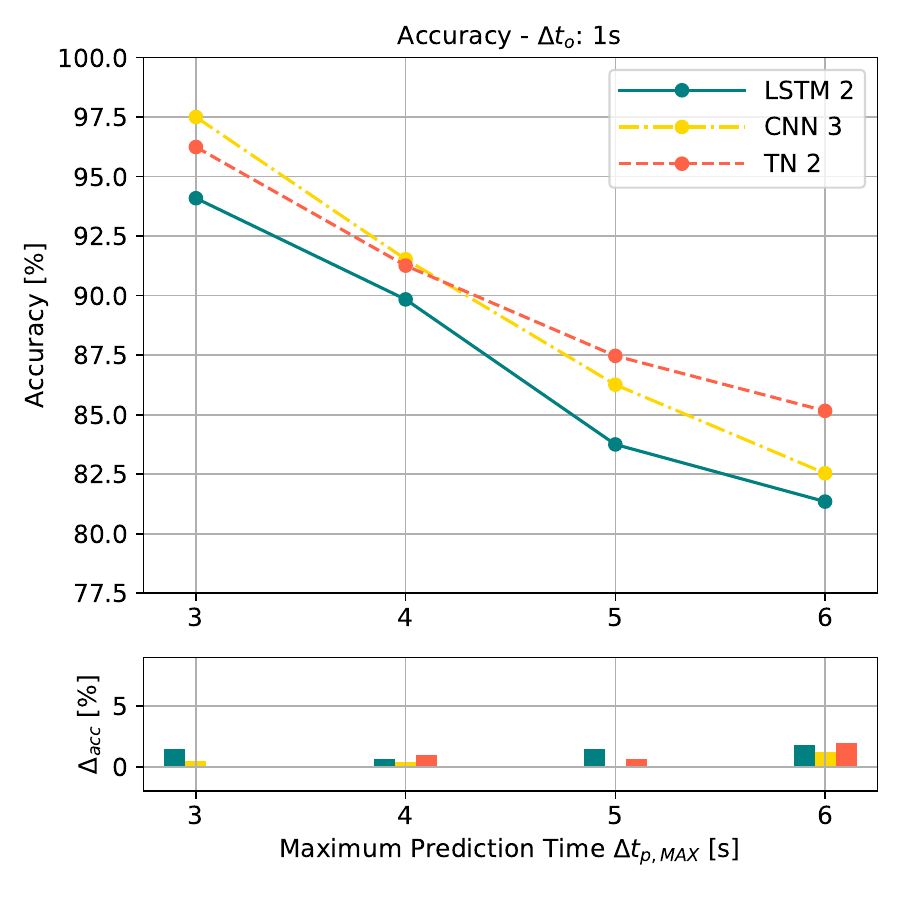}
    \caption{Accuracy for the three selected architectures (LSTM 2, CNN 2, TN 2) for $\Delta t_o=1\textnormal{s}$ and varying values of $\Delta t_{p,\textnormal{MAX}}$ with the corresponding values of $\Delta_{acc}$.}
    \label{fig:acc_1s}
\end{figure}

\begin{figure}
    \centering
    \includegraphics[width=1\linewidth]{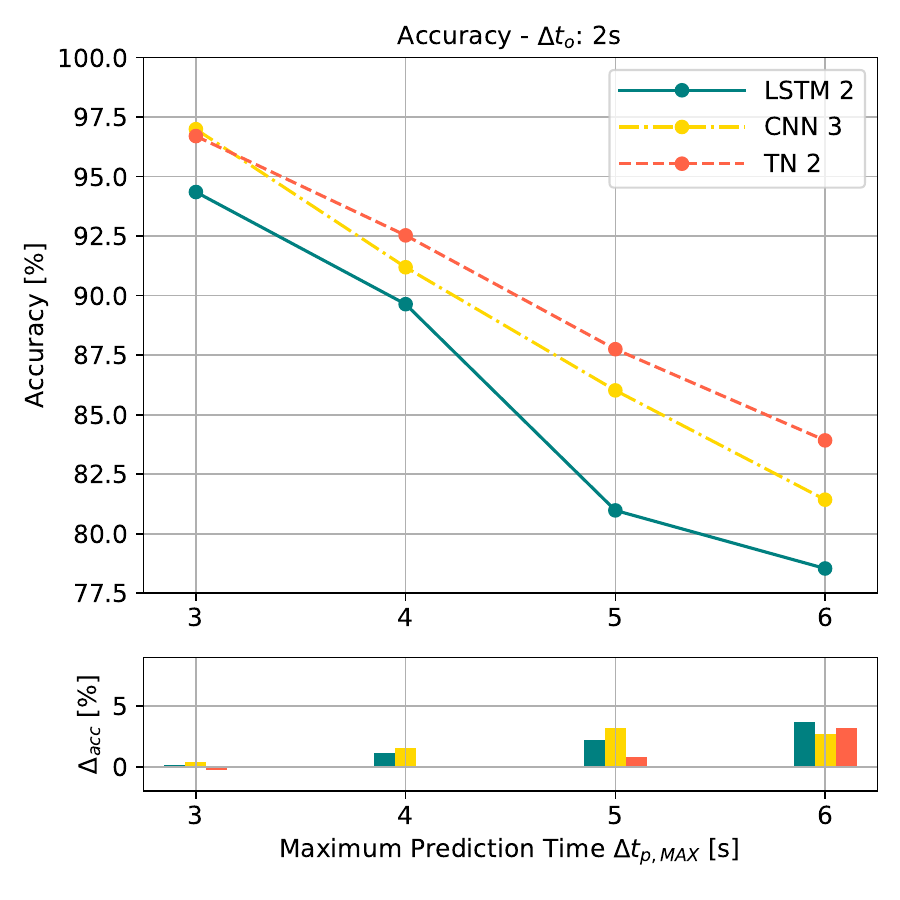}
    \caption{Accuracy for the three selected architectures (LSTM 2, CNN 2, TN 2) for $\Delta t_o=2\textnormal{s}$ and varying values of $\Delta t_{p,\textnormal{MAX}}$ with the corresponding values of $\Delta_{acc}$.}
    \label{fig:acc_2s}
\end{figure}

\begin{figure}
    \centering
    \includegraphics[width=1\linewidth]{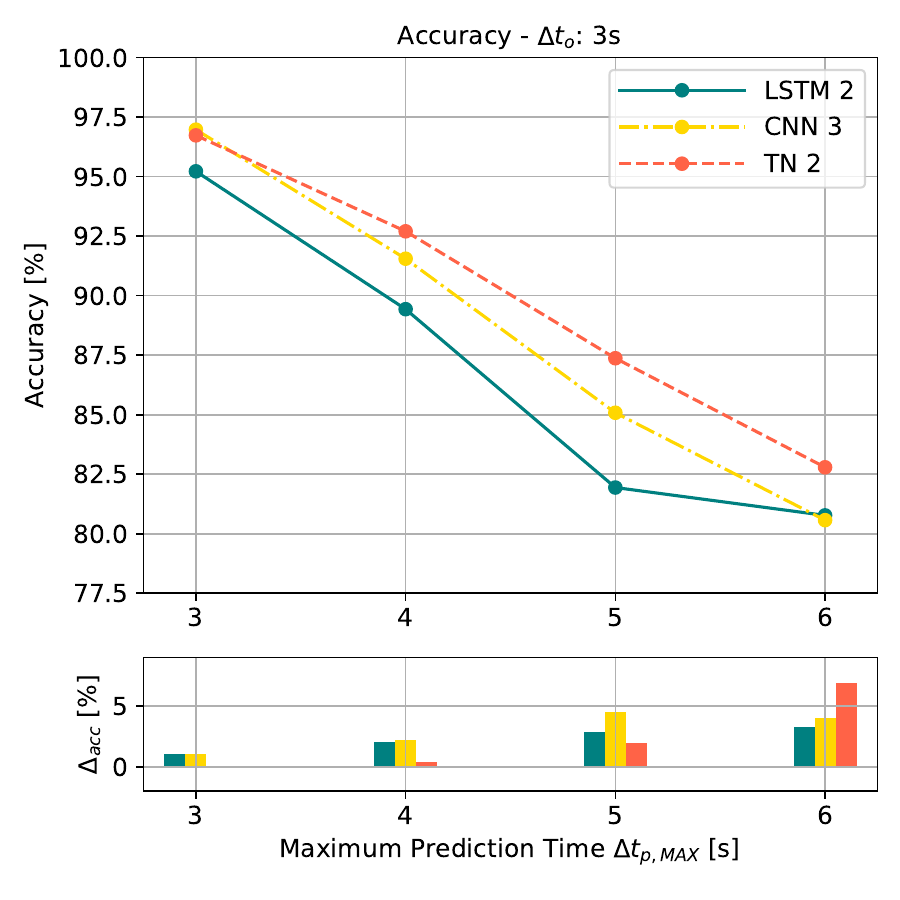}
    \caption{Accuracy for the three selected architectures (LSTM 2, CNN 2, TN 2) for $\Delta t_o=3\textnormal{s}$ and varying values of $\Delta t_{p,\textnormal{MAX}}$ with the corresponding values of $\Delta_{acc}$.}
    \label{fig:acc_3s}
\end{figure}

LSTM 2, CNN 3 and TN 2 were the best performing architectures among each respective class during our tests. While all of them show results which are not dramatically different from each other, LSTM 2 clearly is the one with the worst performances. Fig. \ref{fig:acc_1s}, Fig. \ref{fig:acc_2s} and Fig. \ref{fig:acc_3s} show (respectively for $\Delta t_o=1\textnormal{s}$, $\Delta t_o=2\textnormal{s}$ and $\Delta t_o=3\textnormal{s}$) the accuracy and the $\Delta_{acc}$ for varying values of $\Delta t_{p,\textnormal{MAX}}$. With the exception of the case $(\Delta t_o=3\textnormal{s}, \Delta t_{p,\textnormal{MAX}}=6\textnormal{s})$, in which LSTM 2 and CNN 3 achieve almost identical accuracies and in which LSTM presents a lower $\Delta_{acc}$ with respect to the other two architectures, LSTM 2 has consistently lower accuracies with respect to CNN 3 and TN 2.

On the other hand CNN 3 and TN 2 have more similar and comparable behaviors. With $\Delta t_{p,\textnormal{MAX}}=3\textnormal{s}$, CNN 3 has virtually identical results to TN 2 with slightly higher $\Delta_{acc}$ for $\Delta t_o=2\textnormal{s}$ (see Fig. \ref{fig:acc_2s}) and $\Delta t_o=3\textnormal{s}$ (see Fig. \ref{fig:acc_3s}) while for $\Delta t_o=1\textnormal{s}$ its accuracy is slightly higher than that of TN 2 (see Fig. \ref{fig:acc_1s}). In the case $(\Delta t_o=1\textnormal{s}, \Delta t_{p,\textnormal{MAX}}=4\textnormal{s})$ TN 2 and CNN 3 behave once again pretty similarly (see Fig. \ref{fig:acc_1s}). Finally, in all the other cases, TN 2 outperforms CNN 3 with higher accuracies (although not significantly).

Both networks (TN 2 and CNN 3) show promising results for varying values of $\Delta t_o$ and $\Delta t_{p,\textnormal{MAX}}$ with TN 2 coming out on top in most, but not all, the combinations of observation window and maximum prediction time and showing on average lower over-fitting (expressed by $\Delta_{acc}$).

\begin{figure}
    \centering
    \includegraphics[width=1\linewidth]{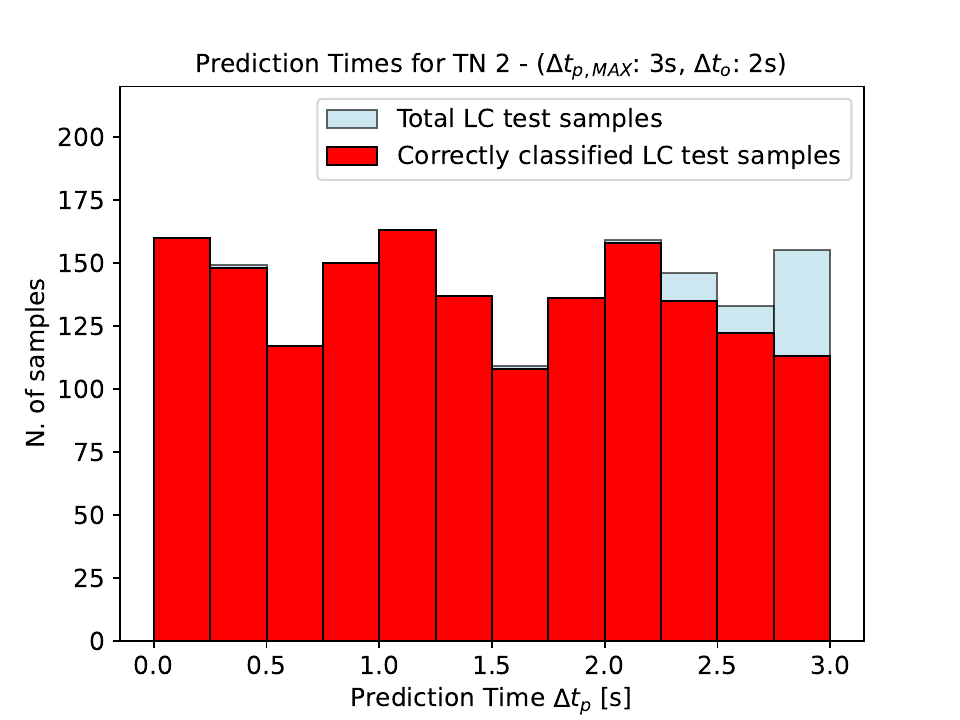}
    \caption{Prediction times of the total LC (LLC and RLC) samples in the ($\Delta t_{p,\textnormal{MAX}}=3\textnormal{s}$, $\Delta t_o=2\textnormal{s}$) configuration and prediction times of the LC samples correctly classified by TN 2 for the same configuration.}
    \label{fig:pw_times_tn_3_2}
\end{figure}

\begin{figure}
    \centering
    \includegraphics[width=1\linewidth]{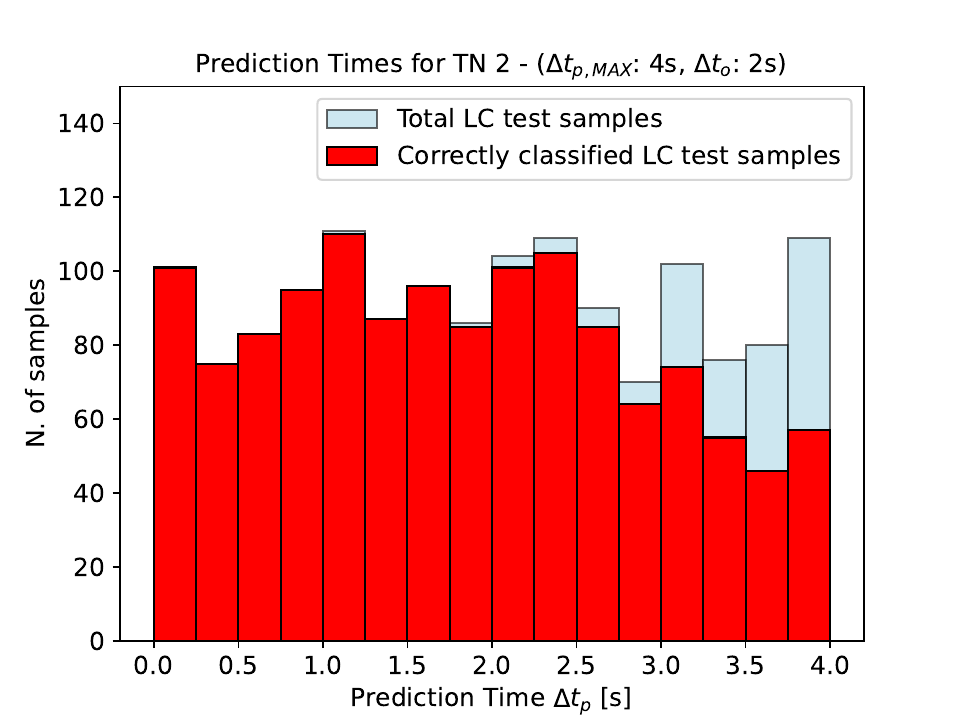}
    \caption{Prediction times of the total LC (LLC and RLC) samples in the ($\Delta t_{p,\textnormal{MAX}}=4\textnormal{s}$, $\Delta t_o=2\textnormal{s}$) configuration and prediction times of the LC samples correctly classified by TN 2 for the same configuration.}
    \label{fig:pw_times_tn_4_2}
\end{figure}

\begin{figure}
    \centering
    \includegraphics[width=1\linewidth]{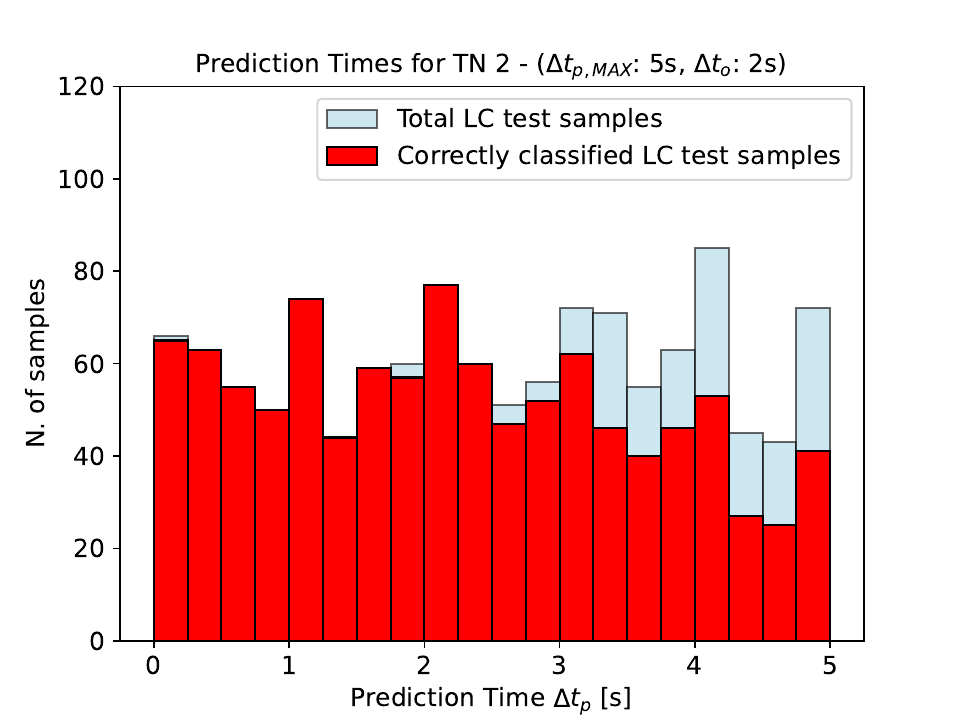}
    \caption{Prediction times of the total LC (LLC and RLC) samples in the ($\Delta t_{p,\textnormal{MAX}}=5\textnormal{s}$, $\Delta t_o=2\textnormal{s}$) configuration and prediction times of the LC samples correctly classified by TN 2 for the same configuration.}
    \label{fig:pw_times_tn_5_2}
\end{figure}

\begin{figure}
    \centering
    \includegraphics[width=1\linewidth]{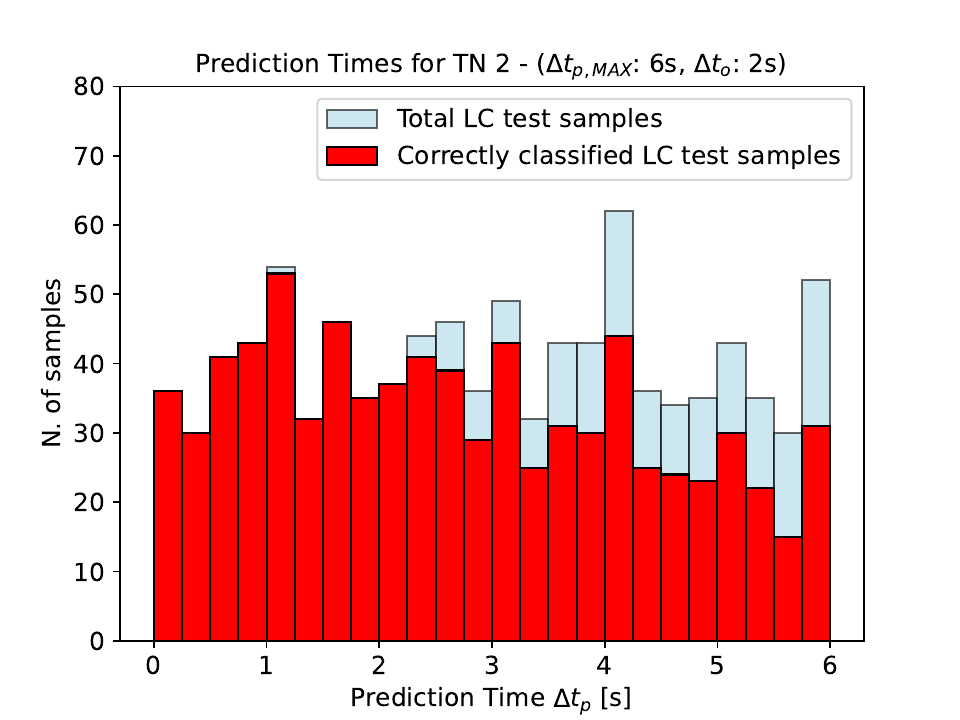}
    \caption{Prediction times of the total LC (LLC and RLC) samples in the ($\Delta t_{p,\textnormal{MAX}}=6\textnormal{s}$, $\Delta t_o=2\textnormal{s}$) configuration and prediction times of the LC samples correctly classified by TN 2 for the same configuration.}
    \label{fig:pw_times_tn_6_2}
\end{figure}

To better visualize the results of TN 2, $\Delta t_o$ was fixed to $2\textnormal{s}$ (an intermediate value) and a comparison was made between the distribution of the prediction times of the LLCs and RLCs which were correctly predicted by TN 2 with the distribution of the total number of LLCs and RLCs included in the test dataset for $\Delta t_{p,\textnormal{MAX}}\in[3\textnormal{s}, 4\textnormal{s}, 5\textnormal{s}, 6\textnormal{s}]$ (respectively Fig. \ref{fig:pw_times_tn_3_2}, Fig. \ref{fig:pw_times_tn_4_2}, Fig. \ref{fig:pw_times_tn_5_2} and Fig. \ref{fig:pw_times_tn_6_2}). It is important to remember why the prediction times are shown as a distribution and not as a fixed value. The initial goal, as stated in Sections \ref{sec:Intro} and \ref{sec:data}, was to predict LCs within a fixed maximum prediction time $\Delta t_{p,\textnormal{MAX}}$. For each LC sample, a prediction time was randomly extracted between $\textnormal{s}$ and $\Delta t_{p,\textnormal{MAX}}$ which explains why prediction times are shown as a distribution.

For all values of $\Delta t_{p,\textnormal{MAX}}$, it appears that the performance of TN 2 started to deteriorate between $2\textnormal{s}$ to $3\textnormal{s}$ before the LC instant (LC samples started to be misclassified) and that training on datasets with higher maximum prediction time did not affect this result (see the point in which the distributions diverge in Fig. \ref{fig:pw_times_tn_3_2}, Fig. \ref{fig:pw_times_tn_4_2}, Fig. \ref{fig:pw_times_tn_5_2} and Fig. \ref{fig:pw_times_tn_6_2}). Although deteriorated, it is interesting that TN 2 is still able to predict a significant portion of LCs even for high prediction times (see Fig. \ref{fig:pw_times_tn_6_2}) showing that the network is not concentrating only on imminent maneuvers. Finally, for each one of Fig. \ref{fig:pw_times_tn_3_2}, Fig. \ref{fig:pw_times_tn_4_2}, Fig. \ref{fig:pw_times_tn_5_2} and Fig. \ref{fig:pw_times_tn_6_2} the relative confusion matrix (which includes also LKs) is reported (respectively Tab. \ref{tab:confusion_tn_3_2}, Tab. \ref{tab:confusion_tn_4_2}, Tab. \ref{tab:confusion_tn_5_2} and Tab. \ref{tab:confusion_tn_6_2}). For each of them most of the misclassifications happen between LCs and LKs and not between LLCs and RLCs, which do not seem to be confused with each other even when training with higher maximum prediction times.

\begin{table}[]
    \centering
    \begin{tabular}{ccc|c|c|c}
        \multicolumn{3}{c}{}&\multicolumn{3}{c}{Predicted} \\ \\
        &&& LK & LLC & RLC  \\ \cline{3-6}
        \multirow{3}{*}{\rotatebox{90}{True}}&&LK & 1607 & 17 & 27 \\ \cline{3-6}
        &&LLC & 28 & 728 & 0 \\ \cline{3-6}
        &&RLC & 39 & 0 & 919 \\ \cline{3-6}
    \end{tabular}
    \caption{Confusion matrix for TN 2 with $\Delta t_{p,\textnormal{MAX}}=3\textnormal{s}$ and $\Delta t_o=2\textnormal{s}$.}
    \label{tab:confusion_tn_3_2}
\end{table}

\begin{table}[]
    \centering
    \begin{tabular}{ccc|c|c|c}
        \multicolumn{3}{c}{}&\multicolumn{3}{c}{Predicted} \\ \\
        &&& LK & LLC & RLC  \\ \cline{3-6}
        \multirow{3}{*}{\rotatebox{90}{True}}&&LK & 1369 & 30 & 32 \\ \cline{3-6}
        &&LLC & 63 & 603 & 0 \\ \cline{3-6}
        &&RLC & 92 & 0 & 716 \\ \cline{3-6}
    \end{tabular}
    \caption{Confusion matrix for TN 2 with $\Delta t_{p,\textnormal{MAX}}=4\textnormal{s}$ and $\Delta t_o=2\textnormal{s}$.}
    \label{tab:confusion_tn_4_2}
\end{table}

\begin{table}[]
    \centering
    \begin{tabular}{ccc|c|c|c}
        \multicolumn{3}{c}{}&\multicolumn{3}{c}{Predicted} \\ \\
        &&& LK & LLC & RLC  \\ \cline{3-6}
        \multirow{3}{*}{\rotatebox{90}{True}}&&LK & 1098 & 43 & 78 \\ \cline{3-6}
        &&LLC & 65 & 486 & 7 \\ \cline{3-6}
        &&RLC & 103 & 3 & 557 \\ \cline{3-6}
    \end{tabular}
    \caption{Confusion matrix for TN 2 with $\Delta t_{p,\textnormal{MAX}}=5\textnormal{s}$ and $\Delta t_o=2\textnormal{s}$.}
    \label{tab:confusion_tn_5_2}
\end{table}

\begin{table}[]
    \centering
    \begin{tabular}{ccc|c|c|c}
        \multicolumn{3}{c}{}&\multicolumn{3}{c}{Predicted} \\ \\
        &&& LK & LLC & RLC  \\ \cline{3-6}
        \multirow{3}{*}{\rotatebox{90}{True}}&&LK & 849 & 50 & 98 \\ \cline{3-6}
        &&LLC & 69 & 398 & 7 \\ \cline{3-6}
        &&RLC & 88 & 5 & 407 \\ \cline{3-6}
    \end{tabular}
    \caption{Confusion matrix for TN 2 with $\Delta t_{p,\textnormal{MAX}}=6\textnormal{s}$ and $\Delta t_o=2\textnormal{s}$.}
    \label{tab:confusion_tn_6_2}
\end{table}

\section{Conclusion}\label{sec:conclusion}
Driverless vehicles at SAE J3016 level 4 are entering commercial business, as demonstrated by public robotaxis services in the USA and China where they are operated in mixed traffic parallel to human-driven vehicles. The driving behavior of L4 vehicles during lane changes is crucial to performing in a safe, comfortable, and energy-efficient manner, especially in complex and dense traffic. Predicting future lane change intentions of human drivers offers the potential for optimized motion planning.

The present paper presents approaches for predicting lane changes based on machine learning techniques. The following approaches were applied in twelve configurations with respect to the prediction time and the observation window: Long Short-Term Memory (LSTM), Convolutional Neural Networks (CNN), and Transformer Networks (TN). In our notation, the number after the used network refers to a specific configuration of hyper-parameters.

Overall, we showed that TN 2 achieved the best results among the tested networks, although CNN 3 came to a close second place and for some configurations of the input dataset got slightly better results than those of the transformer (see Fig. \ref{fig:acc_1s}, Fig. \ref{fig:acc_2s} and Fig. \ref{fig:acc_3s}). This, in our opinion, falls in line with what we observed in the literature: transformer networks are on one hand extremely versatile and effective tools but on the other hand not overwhelmingly superior to other more established algorithms in the field of lane change (LC) intention prediction as of today and with respect to the implementation that we described in this paper.

Regarding the results of the classification problem, we showed that, especially for transformer networks, right lane changes (RLCs) seemed to be harder to predict when taking into account samples with a higher prediction time (see Tab. \ref{tab:lstm_results}, Tab. \ref{tab:cnn_results} and Tab. \ref{tab:trans_results}). This, in our opinion, could be caused either by a difference in the execution times for RLCs and left lane changes (LLCs) or it could be also caused by a difference in traffic density between "fast" and "slow" lanes which affect how, how fast and for how long drivers synchronize their position to an available free slot in an adjacent lane.

Finally, we showed for TN 2 how and when the distribution of the prediction times of the correctly classified LC test samples diverged from that of all the LC test samples. While not perfect, it is interesting to us that TN 2 was still able to make correct predictions at higher prediction times while retaining prediction performances at lower prediction times. Overall, we think that setting maximum prediction time $\Delta t_{p,\textnormal{MAX}}$ to a lower value could be desirable for those situations in which efficiency is of paramount importance, because it would reduce those situations in which a vehicle is erroneously flagged as ready to change lane, leading to unnecessary evasive maneuvers and path recomputations. On the other hand, in those cases in which safety is more important, a higher value for $\Delta t_{p,\textnormal{MAX}}$ could be more desirable. In fact, while unnecessary evasive maneuvers and path recomputations do not necessarily affect safety, some of the LCs could be identified earlier giving more time to the automated vehicle for avoiding a possibly dangerous situation while still guaranteeing a good performance on the imminent ones. If both safety and efficiency are of interest, as it is in most applications, an intermediate setting for $\Delta t_{p,\textnormal{MAX}}$ could be a good compromise. As for the observation window $\Delta t_o$, there was no clear better setting with respect to the accuracy of the prediction and the choice of $\Delta t_o$ should be dependent on that of the desired $\Delta t_{p,\textnormal{MAX}}$.

To conclude, we feel that in the future it would be interesting to explore more in detail the process of hyper-parameter turning, to investigate the reasons of the difference in performance for LLCs and RLCs and finally to test our algorithms on an array of different datasets to better understand how they would react to different situations. We would also like to implement the prediction of lane change intention in L4 vehicle motion planning algorithms to test their performance with respect to safety, comfort and energy efficiency

\section*{Acknowledgments}
The research leading to these results has received funding from the Republic of Austria, Ministry of Climate Action, Environment, Energy, Mobility, Innovation and Technology through grant Nr.~891143 (TRIDENT) managed by the Austrian Research Promotion Agency (FFG). We would like to thank the Science Technology Plan Project of Zhejiang Province (Project Number: 2022C04023) and the Zhejiang Asia-Pacific Intelligent Connected Vehicle Innovation Center Co., Ltd. This work was partially supported by them. A Large Language Model (ChatGPT, OpenAI) was used to assist us, the authors, with writing parts of the Python code which was used to produce the results and figures published in this work.

\newpage

\section{Biography Section}

\begin{IEEEbiography}[{\includegraphics[width=1in,height=1.25in,clip,keepaspectratio]{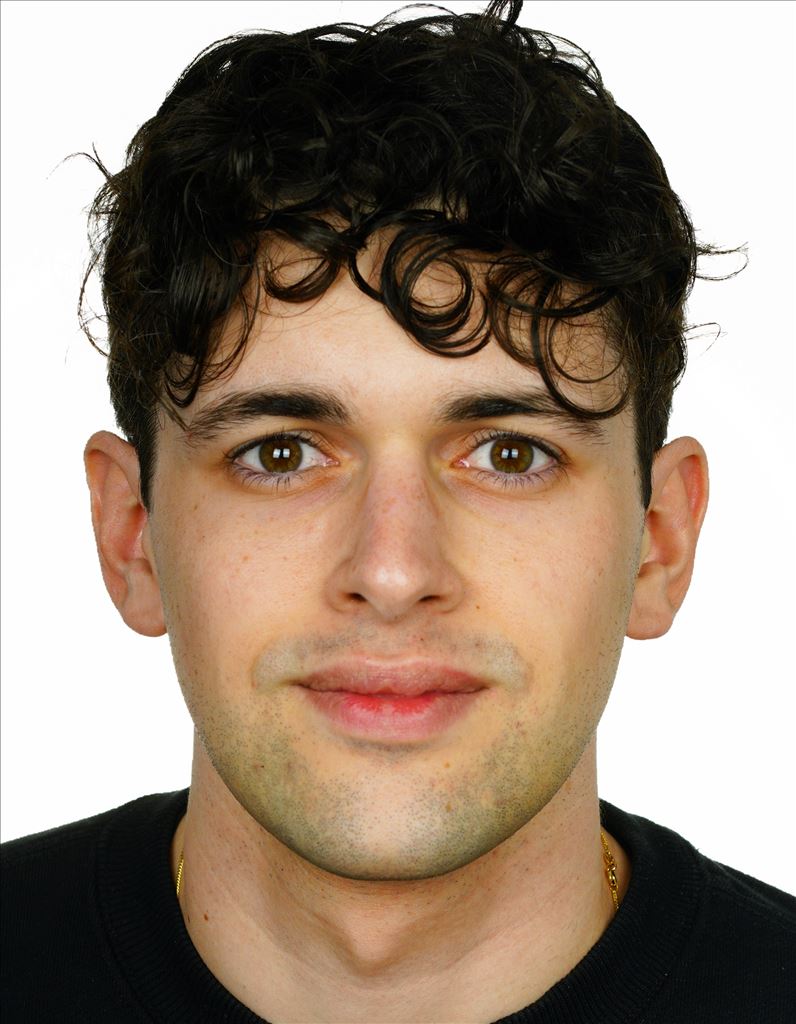}}]{Francesco De Cristofaro}
obtained his BSc in Automation Engineering in 2018 and his MSc in Automation and Control Engineering in 2021 from the Polytechnic University of Milan (Politecnico di Milano). Currently he is pursuing a PhD at the Institute of Automotive Engineering of the Technical University of Graz (Technische Universität Graz) where he was also employed as a university project assistant from 2022 to 2025. His research focuses on prediction algorithms for complex driving scenarios and control architectures for autonomous vehicles.
\end{IEEEbiography}

\begin{IEEEbiography}
[{\includegraphics[width=1in,height=1.25in,clip,keepaspectratio]{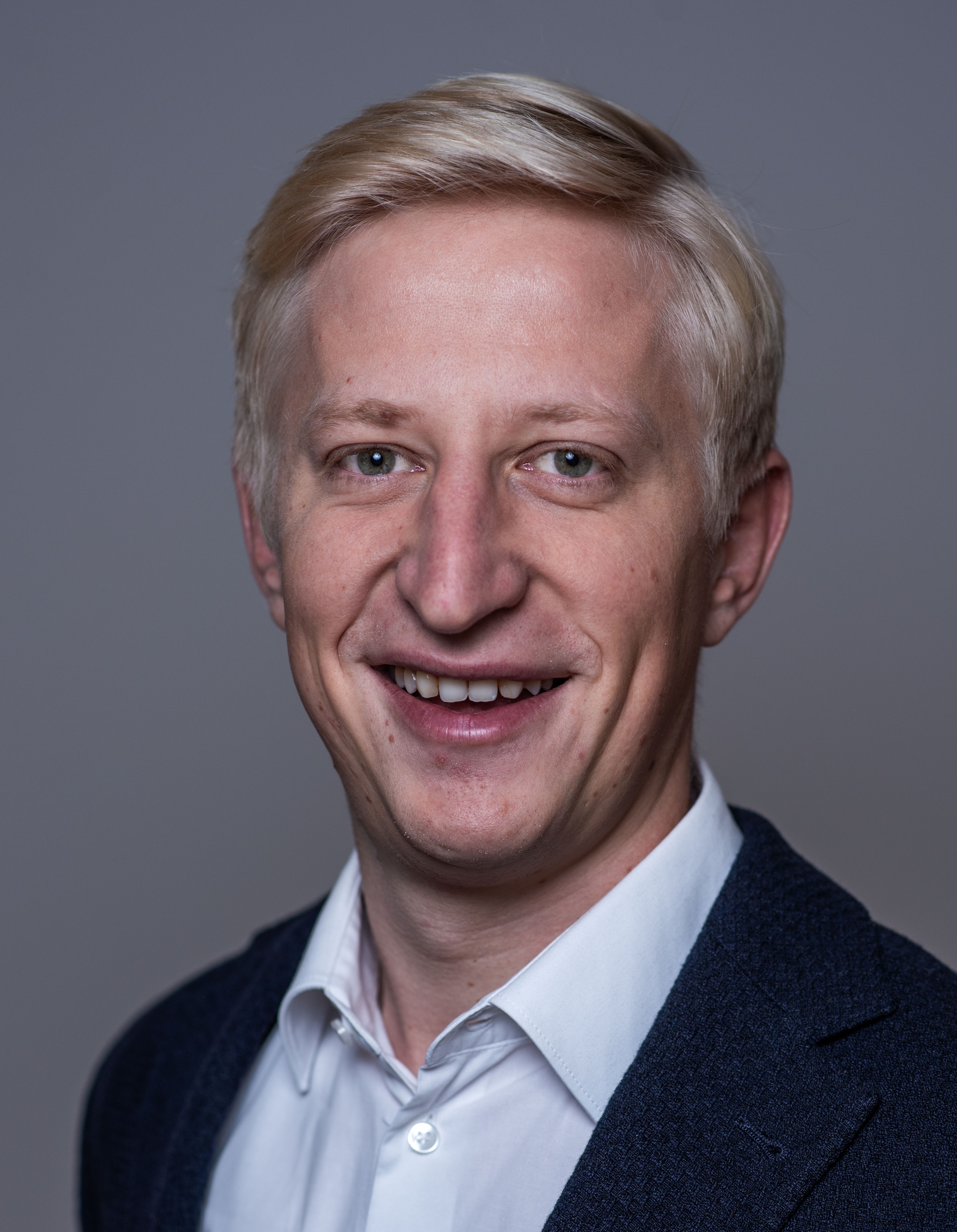}}]{Felix Hofbaur}
holds an MSc in Civil Engineering – Infrastructure from Graz University of Technology. Currently, he is a PhD-candidate at the Institute of Highway Engineering and Transport Planning (Graz University of Technology), where he also works as a university project assistant. His research centers on traffic simulation and modeling of human driver behavior, with a particular emphasis on lane change dynamics.
\end{IEEEbiography}

\begin{IEEEbiography}[{\includegraphics[width=1in,height=1.25in,clip,keepaspectratio]{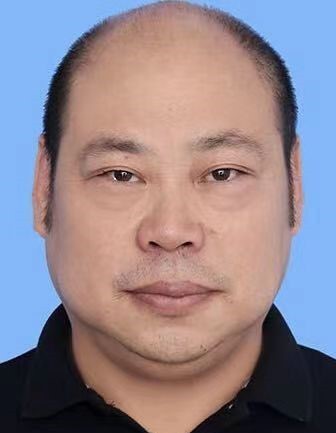}}]{Aixi Yang} is the general Manager of Zhejiang Asia-Pacific Intelligent Connected Vehicle Innovation Center Co., Ltd. Additionally, he serves as a Master's Supervisor at Zhejiang University, holds the title of Senior Engineer, and is recognized as an expert in intelligent manufacturing by the Ministry of Industry and Information Technology of China, as well as an expert in Standard Technology Evaluation. With over 26 years of experience in the automotive industry, he specializes in intelligent chassis and autonomous driving technologies.
\end{IEEEbiography}

\begin{IEEEbiography}[{\includegraphics[width=1in,height=1.25in,clip,keepaspectratio]{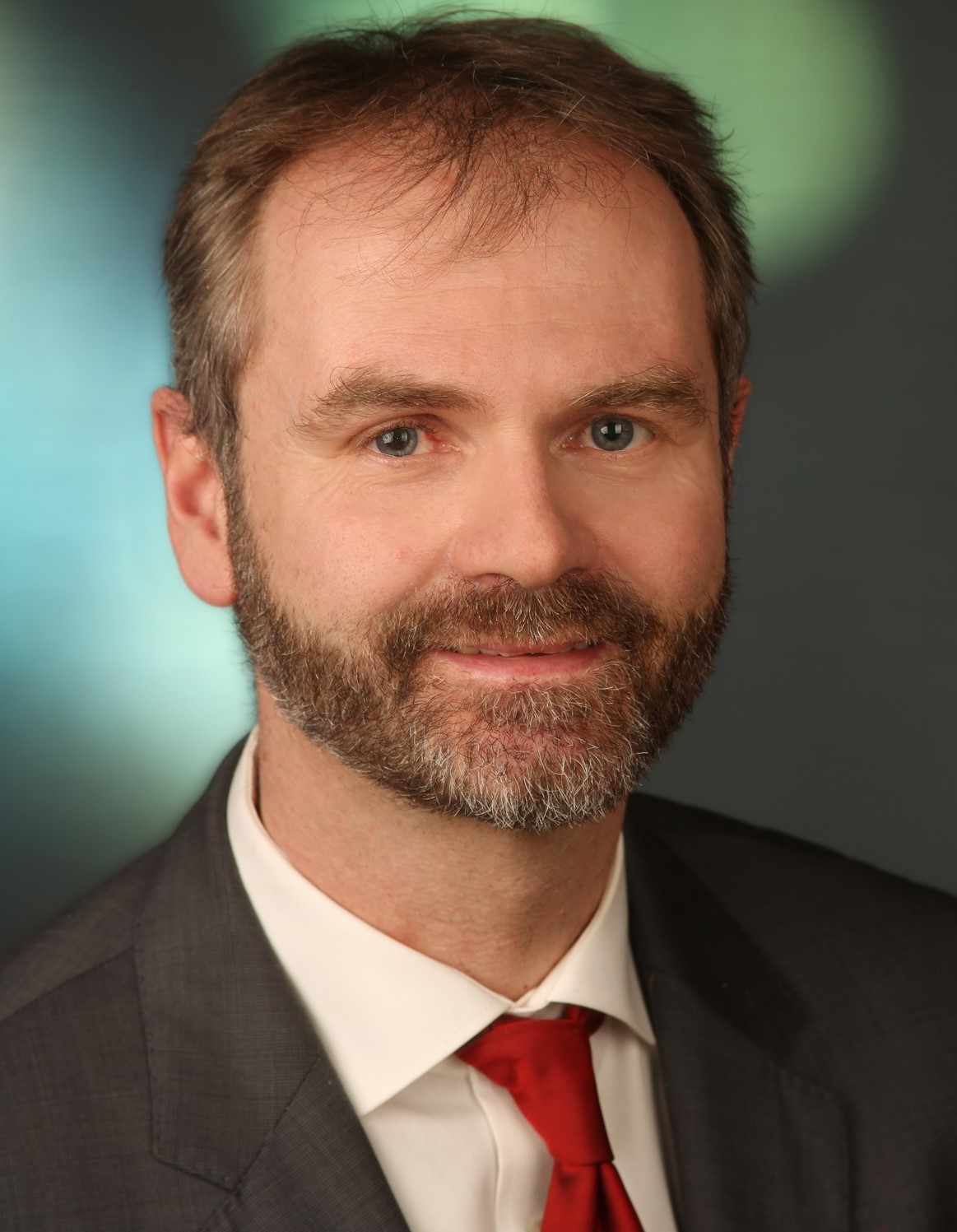}}]{Arno Eichberger} received the degree in mechanical engineering from the University of Technology Graz, in 1995, and the Ph.D. degree in technical sciences, in 1998. From 1998 to 2007, he was employed at MAGNA STEYR Fahrzeugtechnik AG\&Co and dealt with different aspects of active and passive safety. Since 2007, he has been working with the Institute of Automotive Engineering, University of Technology Graz, dealing with automated driving, vehicle dynamics, and suspensions. Since 2012, he has been an Associate Professor holding a venia docendi of automotive engineering.
\end{IEEEbiography}

\vfill

\end{document}